\documentclass{article}

\usepackage{microtype}
\usepackage{graphicx}
\usepackage{subcaption}
\usepackage{booktabs} 

\usepackage{hyperref}

\usepackage[accepted]{icml2026}

\usepackage{amsmath}
\usepackage{amssymb}
\usepackage{mathtools}
\usepackage{amsthm}
\usepackage{ragged2e}
\usepackage{tcolorbox}
\usepackage{listings}
\usepackage{multirow}

\newtcolorbox{AIBox}[2][]{colback=gray!5,colframe=black!50,
    fonttitle=\bfseries,title={#2},#1}

\usepackage[capitalize,noabbrev]{cleveref}

\theoremstyle{plain}

\theoremstyle{definition}

\theoremstyle{remark}

\usepackage[textsize=tiny]{todonotes}

\icmltitlerunning{Strategy Executability in Mathematical Reasoning}

\begin{document}

\twocolumn[
\icmltitle{Strategy Executability in Mathematical Reasoning:
Leveraging Human–Model Differences for Effective Guidance}



  \icmlsetsymbol{equal}{*}

  \begin{icmlauthorlist}
    \icmlauthor{Weida Liang}{nus}
    \icmlauthor{Yiyou Sun}{berkeley}
    \icmlauthor{Shuyuan Nan}{nus}
    \icmlauthor{Chuang Li}{nus}
    \icmlauthor{Dawn Song}{berkeley}
    \icmlauthor{Kenji Kawaguchi}{nus}
  \end{icmlauthorlist}

\icmlaffiliation{nus}{National University of Singapore}
\icmlaffiliation{berkeley}{University of California, Berkeley}

\icmlcorrespondingauthor{Kenji Kawaguchi}{kenji@comp.nus.edu.sg}

  \icmlkeywords{Machine Learning, ICML}

  \vskip 0.3in
]



\printAffiliationsAndNotice{}  

\begin{abstract}
Example-based guidance is widely used to improve mathematical reasoning at inference
time, yet its effectiveness is highly unstable across problems and models—even when the
guidance is correct and problem-relevant.
We show that this instability arises from a previously underexplored gap between
\emph{strategy usage}—whether a reasoning strategy appears in successful solutions—and
\emph{strategy executability}—whether the strategy remains effective when instantiated
as guidance for a target model.
Through a controlled analysis of paired human-written and model-generated solutions, we
identify a systematic dissociation between usage and executability: human- and
model-derived strategies differ in structured, domain-dependent ways, leading to
complementary strengths and consistent source-dependent reversals under guidance.
Building on this diagnosis, we propose \emph{Selective Strategy Retrieval} (SSR), a
test-time framework that explicitly models executability by selectively retrieving and
combining strategies using empirical, multi-route, source-aware signals.
Across multiple mathematical reasoning benchmarks, SSR yields reliable and consistent
improvements over direct solving, in-context learning, and single-source guidance,
improving accuracy by up to $+13$ points on AIME25 and $+5$ points on Apex for compact
reasoning models. Code and benchmark are publicly available at: \url{https://github.com/lwd17/strategy-execute-pipeline}.
\end{abstract}

\section{Introduction}
\label{sec:intro}

Large language models (LLMs) have demonstrated strong performance on mathematical
reasoning tasks, particularly when augmented with inference-time guidance such as worked
examples, concise hints, or high-level reasoning suggestions
\cite{wei2022chain,lewkowycz2022solving,kojima2022large,achiam2023gpt,brown2020language,yao2022react}.
When effective, such guidance does more than add information: it steers the model toward
a particular solution strategy and shapes the sequence of reasoning steps it attempts
to execute.

\begin{figure}[t]
  \vskip 0.1in
  \begin{center}
    \centerline{\includegraphics[width=0.85\linewidth]{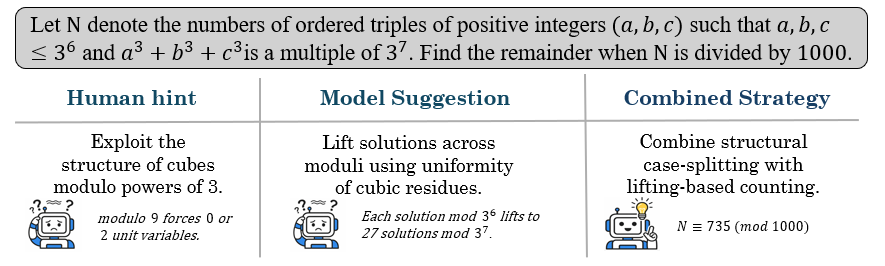}}
    \caption{
      An example illustrating that strategies which appear valid in isolation
      may fail when transferred as guidance.
      In this AIME-level problem, a human-derived structural strategy and a
      model-derived procedural strategy are each insufficient on their own,
      while selectively combining them enables successful execution.
    }
    \label{fig:strategy_source}
  \end{center}
\end{figure}

Despite these gains, guidance-based reasoning remains strikingly unreliable.
Across models and benchmarks, even guidance that is demonstrably correct,
problem-relevant, and extracted from successful solutions often fails to help—and
can sometimes degrade—performance~\cite{madsen2024self,guo2025deepseek}.
These failures recur across domains and model families and cannot be explained by
deficiencies in guidance quality or semantic relevance alone, pointing to a deeper
limitation in how guidance is currently understood.

A central assumption underlying existing approaches is that a reasoning strategy observed
in a successful solution can be reliably carried out when transferred as explicit guidance
to a target model.
In practice, example-based guidance primarily conveys \emph{reasoning strategies}:
high-level decisions about problem decomposition, representation, and solution structure
\cite{simon1971human,chi2006laboratory}.
Human-authored solutions, in particular, often emphasize conceptual insight and global
structure \cite{larkin1980expert,polya1957solve}.
These human strategies are typically concise, abstract, and under-specified, relying on
implicit reasoning steps that may not align with the operational strengths of a target
model.
As a result, the mere presence of a strategy in a correct solution does not ensure
that the target model can effectively use it when prompted.

This gap highlights a distinction that has received little explicit attention:
the difference between whether a strategy appears in a successful solution and whether it
\emph{remains effective as guidance} for a target model.
We refer to the latter as \emph{strategy executability}.
Importantly, executability is assessed operationally---by whether providing the strategy
as guidance under fixed prompting and decoding conditions increases the target model's
likelihood of producing a correct solution, without requiring faithful step-by-step
imitation of the strategy.
This perspective leads to a natural question:
\begin{center}
\emph{Under what conditions does a reasoning strategy remain executable when transferred
as guidance to a target model?}
\end{center}

To address this question, we adopt a strategy-level diagnostic perspective on mathematical
reasoning.
Rather than treating solutions as reasoning traces, we represent each
solution as a composition of high-level strategies.
This abstraction disentangles two notions often conflated in prior work:
\emph{strategy usage}, which captures how frequently a strategy appears in successful
solutions, and \emph{strategy executability}, which reflects whether the strategy remains
effective when instantiated as guidance for a given model.

Using this framework, we analyze paired human-written and model-generated solutions to the
same mathematical problems.
Although both sources often arrive at correct answers, they do so using systematically
different strategies: human solutions rely more on structural insights and conceptual
decompositions, whereas model-generated solutions favor procedural and algebraic
transformations \cite{trinh2024solving,mahdavi2025brains}.
As we show, these differences have concrete consequences for guidance, shaping which
strategies remain executable when transferred.

Figure~\ref{fig:strategy_source} illustrates this phenomenon.
When transferred individually under identical prompting conditions, neither strategy
succeeds; only their selective combination yields an executable reasoning path.
This highlights our central insight: effective guidance depends not on strategy presence
alone, but on executability for the target model.

Motivated by this diagnosis, we propose \textbf{Selective Strategy Retrieval (SSR)}, a
lightweight inference-time framework that explicitly models strategy executability.
SSR selectively retrieves strategies from human-written and model-generated
solutions based on empirical executability signals.
It operates purely at test time and requires no modification to the underlying model,
training data, or decoding procedure.

\begin{figure}[t]
  \vskip 0.1in
  \begin{center}
    \centerline{\includegraphics[width=0.85\linewidth]{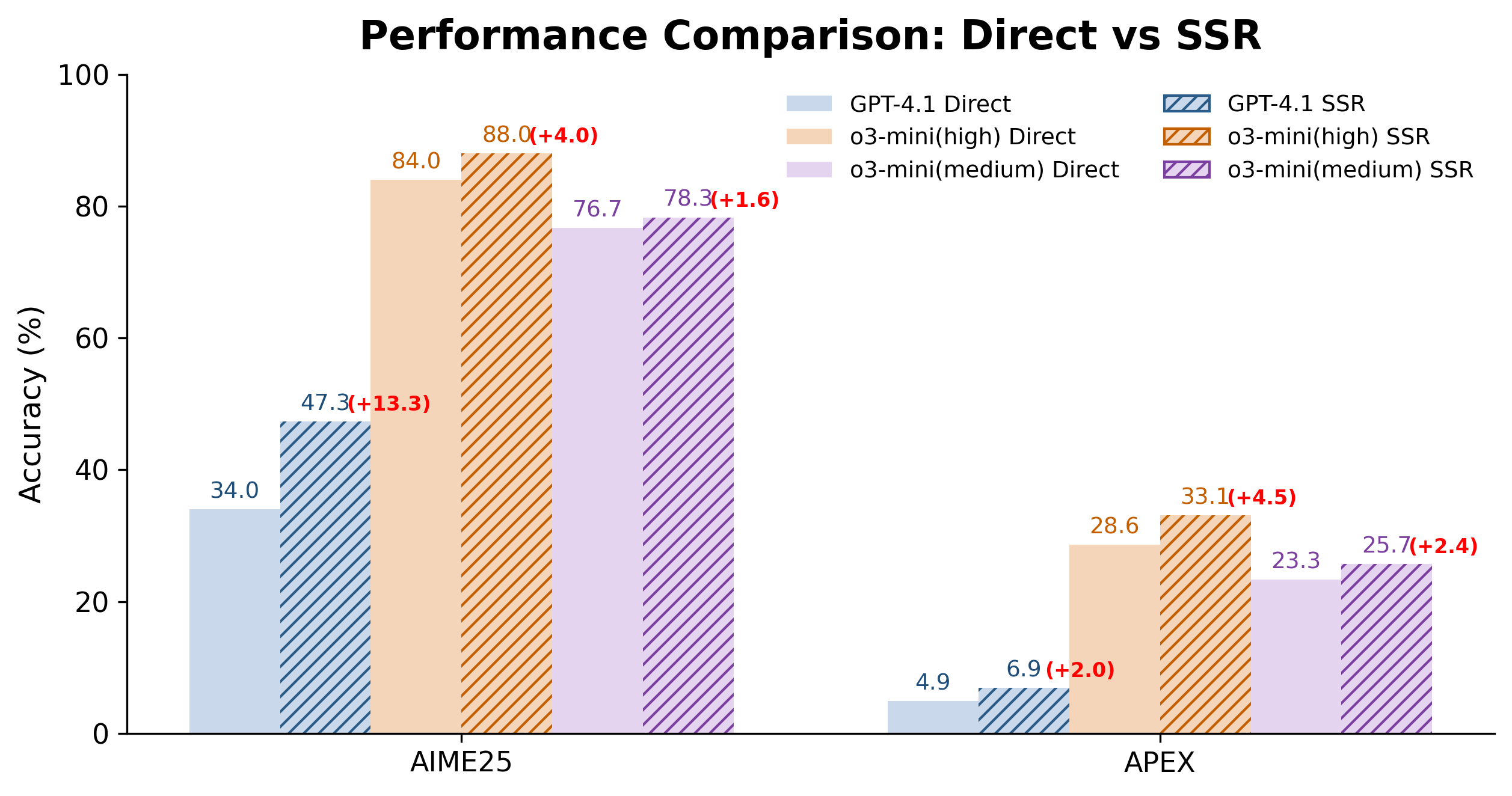}}
    \caption{
      Performance gains from Selective Strategy Retrieval (SSR) on closed-source
      reasoning models (GPT-4.1 and o3-mini), measured by pass@1 and averaged
      over five runs.
    }
    \label{fig:ssr_performance}
  \end{center}
\end{figure}

Empirically, SSR yields consistent improvements across open-source and closed-source
reasoning models.
On closed-source models, SSR improves accuracy over direct prompting by approximately
\(\mathbf{+4\sim13}\) points on AIME25 and \(\mathbf{+2\sim5}\) points on Apex for GPT-4.1 and
o3-mini (Figure~\ref{fig:ssr_performance}), demonstrating that explicitly modeling strategy
executability is key to robust reasoning gains.

Our contributions are summarized as follows:
\begin{itemize}
\item We identify a systematic dissociation between strategy usage and
executability in mathematical reasoning.

\item To enable controlled analysis of this dissociation, we construct
HM-ReasoningBench, a paired dataset of competition-level problems with
human-written and model-generated solutions.

\item Building on this diagnosis, we propose Selective Strategy Retrieval (SSR),
a test-time framework that operationalizes strategy executability through
selective strategy combination.

\item We demonstrate that explicitly modeling strategy executability—rather than
strategy prevalence or semantic relevance—leads to robust improvements across
multiple mathematical benchmarks.
\end{itemize}

\paragraph{Conflict of Interest Disclosure.}
The authors declare no financial or substantive conflicts of interest related to this work.

\section{Related Works}
\label{sec:related}

\noindent\textbf{Example-Based Reasoning Guidance.}
Inference-time guidance, such as worked examples, reasoning traces, or high-level hints,
is widely used to improve reasoning in large language models
\cite{brown2020language,wei2022chain,yao2022react,kojima2022large,liu2022generated,zhou2022least,
rubin2022learning,shum2023automatic,wu2023self,fernando2023promptbreeder,diao2024active,
zhang2025booststep,cao2025step}.
While effective in some cases, prior work largely assumes that guidance is transferable
across models and contexts, typically selecting examples based on semantic similarity or
correctness \cite{zelikman2022star,yao2023tree}.
Recent studies show that additional guidance can be unreliable and may even degrade
performance for certain models or tasks \cite{madaan2023self,guo2025deepseek}, suggesting
that the key challenge lies in whether guidance is executable by the target model.

\noindent\textbf{Reasoning Traces and Strategy Abstraction.}
A large body of work represents solutions as step-by-step reasoning traces for supervision,
explanation, or iterative refinement
\cite{cobbe2021training,wang2022self,zelikman2022star,chowdhury2025zero,mukherjee2025premise,jiang2025makes}.
However, recent work questions the faithfulness of such traces, noting that they may be
post hoc or weakly coupled with model decision-making
\cite{creswell2022faithful,xu2024pride,wu2025more,munkhbat2025self}.
This has motivated abstractions toward higher-level reasoning structure
\cite{yu2025chain}, as well as process-level methods such as process reward models
\cite{hu2025coarse,younsi2025accurate}, which typically operate during training or decoding.

Related approaches introduce explicit strategy- or plan-level control, such as routing
problems to strategies or selecting plans prior to generation
\cite{xu2025teaching,qi2025plan}, but do not analyze whether such strategies remain
executable across models or contexts.

In contrast, we treat reasoning strategies as analytical objects.
We abstract strategies from human-written and model-generated solutions and study
\emph{strategy executability}—whether a strategy that appears in a given solution
can be operationalized as guidance—revealing a systematic gap between strategy prevalence
and effectiveness.
This perspective motivates selective strategy retrieval based on empirical executability
signals and human--model differences.

\label{sec:analysis}
\section{Strategy-Level Differences Between Human and Model Solutions}
\label{sec:strategy_analysis}

Before assessing whether a reasoning strategy can serve as effective guidance, we examine
how strategies are employed by different solvers.
Although human-written and model-generated solutions often reach correct answers on the
same problems, they do so through systematically different strategic choices.
This section provides a strategy-level analysis of these differences across problem
domains and establishes the empirical foundation for the executability study in
Section~\ref{subsec:strategy_executability}.

\begin{figure*}[t]
  \begin{center}
    \centerline{\includegraphics[width=0.85\linewidth]{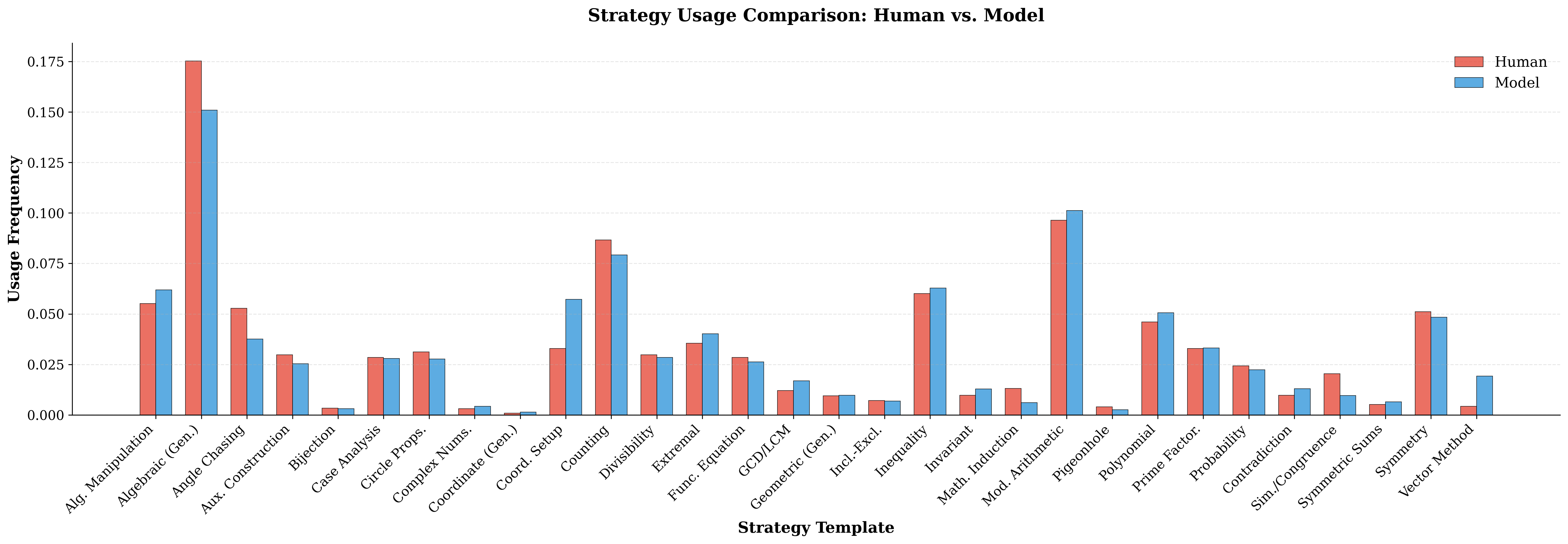}}
    \caption{
      Normalized strategy usage in human-written and model-generated solutions,
      aggregated across problems with per-problem normalization.
      For each problem, strategies contribute equally, ensuring
      that multi-strategy solutions do not dominate the statistics.
    }
    \label{fig:strategy_dominance}
  \end{center}
\end{figure*}

\subsection{Dataset and Paired Solution Setting}
\label{subsec:dataset}

We conduct our analysis on \textbf{HM-ReasoningBench}, which contains $4{,}895$ challenging
mathematical problems, each paired with a human-written solution and a model-generated solution.
Problems are drawn from Omni-Math~\cite{gao2024omni} and HARP~\cite{yue2024harp}, with HARP
restricted to difficulty level $\geq 6$.
The dataset spans algebra, geometry, number theory, combinatorics, and mixed-topic
problems; additional statistics are reported in Appendix~\ref{app:dataset}.

\subsection{Strategy Abstraction}
\label{subsec:extract_strategy}

To enable strategy-level comparison, we represent each solution as a small set of
high-level reasoning strategies.
Each solution is associated with multiple strategies (typically 3--5), reflecting the
compositional nature of non-trivial mathematical reasoning. Strategies are treated as unordered, non-exclusive attributes of a solution, and the
analysis in this section concerns which strategies appear rather than how they are
executed.

Strategies are extracted using a prompting pipeline designed to identify
\emph{transferable reasoning patterns} that generalize beyond individual problems.
Extracted strategies are mapped via rule-based matching to a predefined library of
30 canonical strategy templates, defined based on standard competition guidebooks and
canonical treatments of mathematical problem solving
\cite{polya1957solve,engel1998problem,zeitz2016art}.
Full prompt details and strategy category definitions are provided in
Appendix~\ref{app:sa_prompt} and Appendix~\ref{app:category}.

To ensure comparability across problems, we apply per-problem normalization when
aggregating statistics.

\subsection{Strategy Usage Differences}
\label{subsec:global_strategy}

We first compare strategy usage between human-written and model-generated solutions,
both in aggregate and conditioned on problem domain.

\noindent\textbf{Global preferences.}
Aggregated across all problems, human and model solutions exhibit clear but moderate
differences in their overall strategy distributions.
As shown in Figure~\ref{fig:strategy_dominance}, human-written solutions place greater
emphasis on geometry- and structure-oriented strategies, including auxiliary
constructions, symmetry, angle chasing, and invariant-based reasoning.
Model-generated solutions, in contrast, rely more heavily on algebraic manipulations,
coordinate formulations, and equation-driven transformations.

These trends align with established observations in mathematical problem solving:
expert humans favor relational and structural abstractions, whereas contemporary
reasoning models more often adopt procedural strategies that decompose problems into
explicit symbolic operations~\cite{ruis2411procedural,trinh2024solving}.

\begin{figure*}[t]
  \vskip 0.1in
  \begin{center}
    \centerline{\includegraphics[width=0.9\linewidth]{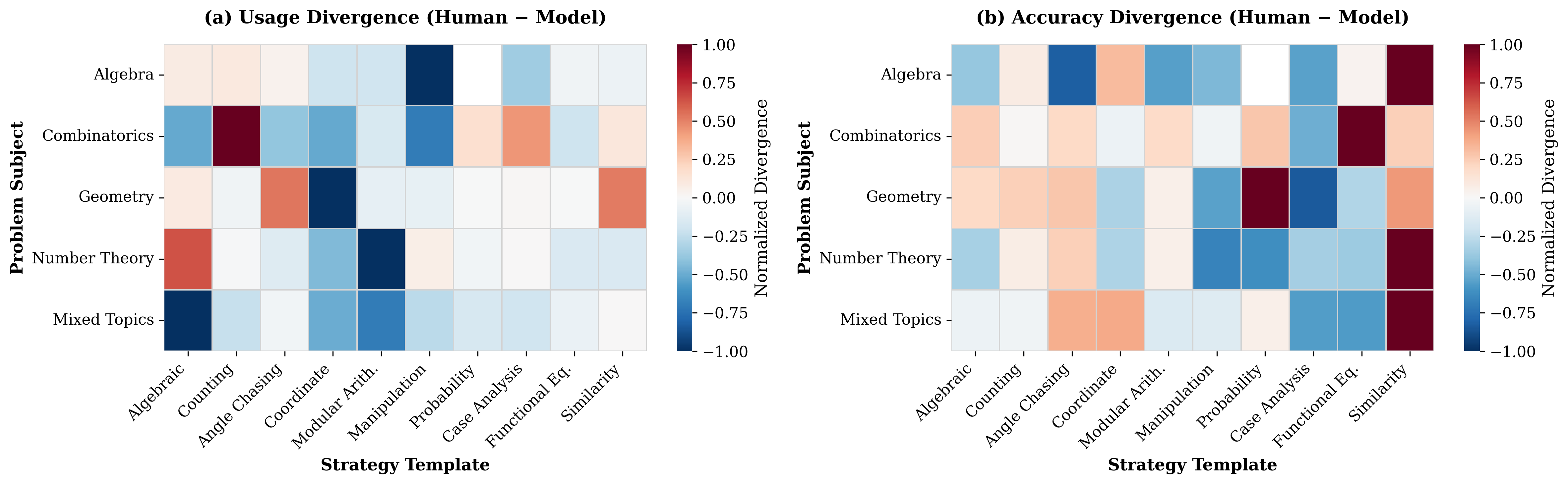}}
    \caption{
      Strategy-level divergence between human-written and model-generated solutions.
      (a) Normalized differences in strategy usage.
      (b) Normalized differences in strategy-guided accuracy.
    }
    \label{fig:usage_accuracy_divergence}
  \end{center}
\end{figure*}

\noindent\textbf{Domain-conditioned divergence.}
Conditioning on problem subject reveals substantially sharper differences.
Geometry exhibits the largest divergence: human solutions strongly favor
construction- and relation-driven strategies, while model solutions disproportionately
adopt coordinate-based reductions
(Figure~\ref{fig:usage_accuracy_divergence}(a)).
In contrast, algebra and number theory show much closer alignment in strategy usage,
likely reflecting their more uniformly symbolic structure.
Additional examples are provided in Appendix~\ref{app:case_studies}.

These usage results should be interpreted as \emph{descriptive} evidence of structured
source differences rather than as direct evidence of guidance quality.
They motivate the executability analysis below: if human-written and model-generated
solutions organize reasoning differently, then the strategies they provide as guidance
may also transfer differently.

\begin{tcolorbox}[colback=gray!5,colframe=gray!40,title=\textbf{Takeaway},boxrule=0.5pt]
Human-written and model-generated solutions differ systematically in the strategies they
employ, with divergences that are strongly structured by problem domain.
These usage patterns characterize how solutions are constructed, but do not indicate
whether a strategy is reliable when transferred as guidance.
\end{tcolorbox}

\subsection{Strategy-Guided Accuracy Divergence}
\label{subsec:strategy_executability}

We now evaluate \emph{strategy executability}: whether an individual strategy extracted
from a solution can be operationalized by a target model when provided as
explicit guidance.

\noindent\textbf{Setup.}
We evaluate two compact reasoning models, \textbf{Qwen3-8B} and
\textbf{DeepSeek-R1-Distill-Qwen-7B}.
Each solution typically contains multiple extracted strategies.
Rather than selecting a representative or random strategy, we treat each extracted
strategy as an independent evaluation unit.
For a given problem--strategy pair, the strategy is provided alone as guidance under a
fixed prompting and decoding protocol, and effectiveness is measured by final answer
correctness.
Results are aggregated at the strategy level; unless otherwise stated, each strategy is
evaluated with multiple decoding trials and averaged to account for stochasticity.
Both models exhibit consistent qualitative trends, so results are aggregated.

\noindent\textbf{Results.}
Figure~\ref{fig:usage_accuracy_divergence}(b) reveals a clear dissociation between
strategy usage and strategy executability.
Strategies that are frequently used by a source solver do not necessarily yield higher
accuracy when transferred as guidance.

Procedural strategies—such as \emph{case analysis} and \emph{coordinate setups}—are often
more executable when sourced from model-generated solutions, particularly in geometry
and mixed-topic problems.
Conversely, structurally grounded strategies—such as
\emph{similarity/congruence} and \emph{prime factorization}—transfer more
reliably when derived from human solutions, despite being less prevalent in model usage.

One useful interpretation is that executability depends on two task properties:
\emph{abstraction leverage} (whether a high-level reformulation immediately reduces the
problem) and \emph{execution load} (whether success requires many explicit intermediate
operations). Structural guidance tends to help more when abstraction leverage is high,
whereas procedural guidance becomes more reliable when execution load dominates.
We return to this task-conditional pattern in Section~\ref{subsec:human_analysis} and
Appendix~\ref{app:style_pool}.



\section{Selective Strategy Retrieval}
\label{sec:ssr}

The analysis in Section~\ref{subsec:strategy_executability} shows that naive strategy reuse
fails for systematic reasons.
In particular, using strategy frequency as a proxy for executability is unreliable;
semantic relevance alone does not ensure operational alignment; and committing to a
single strategy source ignores strong, domain-dependent reversals.
Together, these observations rule out retrieval schemes based solely on usage statistics,
surface similarity, or uniform source preference.
We therefore treat \emph{strategy executability} as the primary criterion for strategy
selection.

We introduce \textbf{Selective Strategy Retrieval (SSR)}, a test-time framework that
selects strategies using source-dependent and context-conditioned executability signals
and provides up to five strategies as guidance per problem.

\subsection{Strategy Knowledge Graph}

Executability is inherently relational: whether a strategy is executable depends on its
interaction with problem structure, reasoning category, and source. 
We therefore organize reasoning knowledge from HM-ReasoningBench into a heterogeneous
graph $\mathcal{G} = (V, E)$ with nodes corresponding to \textbf{problems} $V_p$,
\textbf{strategies} $V_s$, and \textbf{categories} $V_c$.
Edges encode observed problem--strategy usage and category membership.
This representation allows executability signals to propagate across related problems
while preserving the category-level regularities identified in
Section~\ref{sec:strategy_analysis}.

\noindent\textbf{Source-aware retention.}
Because executability depends strongly on strategy source, SSR does not treat all
strategies within a category as equally reliable.
For each category and strategy type, we preferentially retain strategies from the source
(human or model) that exhibits higher empirical executability under guidance.
This design retains complementary strategies
when coverage is sparse.

\noindent\textbf{Graph representation learning.}
To encode executability patterns, we learn structure-aware node embeddings over
$\mathcal{G}$ using a heterogeneous graph neural network with transformer-style message
passing.
The model is trained with a contrastive objective that separates successful from
unsuccessful problem--strategy pairings.
The learned embeddings encode empirical signals of strategy executability
(Appendix~\ref{app:gnn-config}), which serve as structural features for downstream executability prediction,
as described in Section~\ref{subsec:executability_model}.

\begin{figure*}[t]
  \begin{center}
    \centerline{\includegraphics[width=0.8\linewidth]{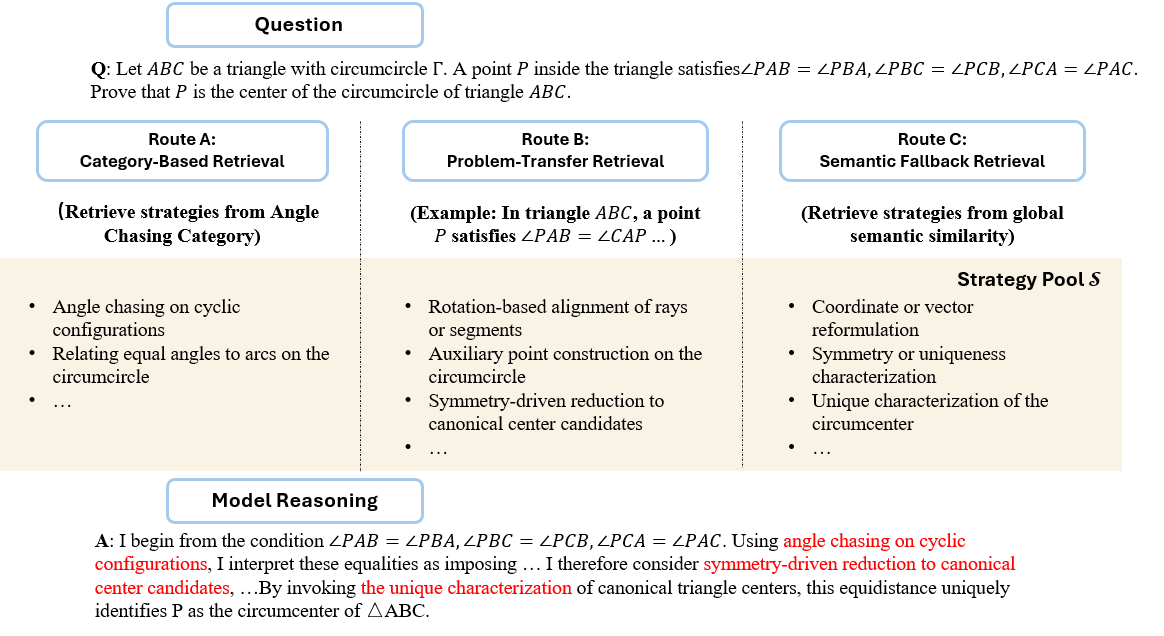}}
    \caption{
      Multi-route strategy retrieval in Selective Strategy Retrieval (SSR).
      Complementary retrieval routes capture category-level regularities,
      problem-specific transfer, and semantic coverage, forming the candidate
      set $\mathcal{S}(x)$.
    }
    \label{fig:multi_route_retrieval}
  \end{center}
\end{figure*}

\subsection{Problem Representation}

At test time, executability must be inferred for a new problem $x$ without direct
supervision.
We first embed $x$ using a pretrained sentence encoder to identify a neighborhood
$\mathcal{N}(x)$ of semantically related training problems.

Rather than retrieving strategies by surface similarity, SSR aggregates the
graph embeddings of problems in $\mathcal{N}(x)$ to construct a transferred
representation $h_x$.
Neighbors are weighted by semantic similarity via a temperature-scaled softmax, allowing
relevant contexts to dominate.
This representation provides a structure-aware abstraction of $x$, enabling retrieval
based on learned executability patterns rather than semantic overlap alone.

\subsection{Multi-Route Strategy Retrieval}

No single notion of relevance reliably predicts executability.
SSR therefore retrieves candidate strategies through three complementary routes, whose
union forms the candidate set $\mathcal{S}(x)$
(Appendix~\ref{app:pool_implementation}).

\noindent\textbf{Route A: Category-conditioned retrieval.}
SSR retrieves strategies retained for categories compatible with $h_x$, capturing
coarse-grained but robust executability signals that generalize across problems.

\noindent\textbf{Route B: Problem-transfer retrieval.}
To capture fine-grained, context-dependent executability, SSR retrieves strategies that
were empirically effective when guiding solutions to problems in $\mathcal{N}(x)$.

\noindent\textbf{Route C: Semantic fallback retrieval.}
When executability evidence is sparse, SSR retrieves a small number of semantically
similar strategies as fallback, ensuring coverage without assuming executability from
similarity.

\subsection{Modeling Strategy Executability}
\label{subsec:executability_model}

The multi-route retrieval step produces a diverse but over-complete candidate set
$\mathcal{S}(x)$, within which only a subset of strategies are expected to be executable
for the target model. Given a problem $x$ and a candidate strategy $s \in \mathcal{S}(x)$, SSR aims to estimate
the utility of providing $s$ as inference-time guidance to a target reasoning model.
We formalize it as a model-relative, protocol-relative quantity:
\begin{equation}
U(s \mid x; m, \pi)
\;=\;
\mathbb{P}\bigl(\text{success}=1 \;\big|\; x, s, m, \pi \bigr),
\label{eq:executability_def}
\end{equation}
where $m$ denotes the target model and $\pi$ denotes a fixed prompting and decoding
protocol (including prompt template, temperature, and context budget).
Intuitively, $U(s \mid x; m, \pi)$ captures the probability that providing strategy $s$
as guidance enables model $m$ to produce a correct solution for problem $x$ under
controlled inference conditions. Success refers to pass@1 unless otherwise stated.

\noindent\textbf{Empirical supervision.}
The executability utility in Eq.~\eqref{eq:executability_def} is not directly observable.
To obtain supervision, we evaluate strategy-guided execution outcomes on a training
split of HM-ReasoningBench.
For each problem--strategy pair $(x,s)$, we run the target model $m$ under protocol $\pi$
for $T$ independent decoding trials and record binary outcomes
$y_{x,s,t} \in \{0,1\}$ indicating whether the final answer is correct.
We treat these outcomes as Bernoulli samples from an underlying success probability
$p_{x,s}$ and estimate a posterior mean executability score using a Beta--Binomial model:
\begin{equation}
\tilde U(s \mid x)
\;=\;
\mathbb{E}[p_{x,s} \mid y_{x,s,1:T}]
\;=\;
\frac{\alpha + \sum_{t=1}^{T} y_{x,s,t}}{\alpha + \beta + T},
\label{eq:beta_binomial}
\end{equation}
with a weakly informative prior $(\alpha,\beta)$.
In all main experiments, we use the symmetric choice $\alpha=\beta=1$;
sensitivity to other weak priors is reported in Appendix~\ref{app:prior_sensitivity}.
This formulation explicitly accounts for decoding stochasticity and yields a calibrated
estimate of strategy executability.

\noindent\textbf{Executability predictor.}
To generalize beyond observed pairs and enable efficient ranking at test time, we learn
a parametric estimator $\hat U_\theta(s \mid x)$ that predicts executability from
problem--strategy features.
We construct a feature representation $\phi(x,s)$ that aggregates complementary signals,
including:
(i) semantic alignment between $x$ and $s$,
(ii) structural proximity derived from the strategy knowledge graph,
and (iii) route- and source-specific indicators reflecting how $s$ was retrieved.
The executability predictor is defined as
\begin{equation}
\hat U_\theta(s \mid x)
\;=\;
\sigma\!\left( \theta^\top \phi(x,s) \right),
\label{eq:utility_model}
\end{equation}
where $\sigma$ is the logistic function.

We train $\hat U_\theta$ using trial-level supervision by minimizing the negative
log-likelihood of observed outcomes:
\begin{equation}
\begin{aligned}
\mathcal{L}(\theta)
&=
-\!\!\sum_{(x,s)} \sum_{t=1}^{T}
\Bigl[
y_{x,s,t} \log \hat U_\theta(s \mid x)
\\
&\qquad\qquad
+
(1-y_{x,s,t}) \log \bigl(1-\hat U_\theta(s \mid x)\bigr)
\Bigr].
\end{aligned}
\label{eq:utility_loss}
\end{equation}
with $\ell_2$ regularization on $\theta$.
This objective encourages $\hat U_\theta$ to approximate the true executability
probability in Eq.~\eqref{eq:executability_def}.

To avoid notation ambiguity, we use three distinct quantities throughout:
$y_{x,s,t}$ denotes a trial-level execution outcome,
$\tilde U(s\mid x)$ denotes the aggregated posterior executability target,
and $\hat U_\theta(s\mid x)$ denotes the predictor output.
The graph encoder and the executability predictor are also trained in two distinct
stages: the graph model uses thresholded $\tilde U$ only to shape structural
representations, while the final predictor is fit with trial-level outcomes.
Appendix~\ref{app:gnn-config} and Appendix~\ref{app:predictor_ablation} provide the
corresponding implementation details and ablations.

The role of the graph model is thus to provide structure-aware representations,
while executability estimation and ranking are handled by a separate supervised predictor. 
\noindent\textbf{Calibration and ranking.}
Because $\hat U_\theta$ is used for cross-route and cross-source comparison, we apply
temperature scaling on a held-out validation set to calibrate predicted probabilities.
At inference time, SSR ranks candidate strategies for problem $x$ by their calibrated
utility scores and selects a small subset with highest estimated executability.

\subsection{Using Strategies as Guidance}
\label{subsec:strategy_guidance}

SSR outputs a small set of abstract strategy hints describing general reasoning
approaches rather than concrete solution steps.
At the strategy level, SSR preserves flexibility while aligning guidance
with the operational strengths of the target model.
The prompting format is shown in Appendix~\ref{app:sg_prompt}.
\begin{table*}[t]
  \caption{
    Accuracy (\%) comparison between Selective Strategy Retrieval (SSR),
    single-source guidance (H/M), in-context learning (ICL),
    and direct solving (DS) across three benchmarks.
    Best results are shown in bold.
  }
  \label{tab:main_results}
  \begin{center}
    \begingroup
    \scriptsize
    \setlength{\tabcolsep}{3pt}
    \renewcommand{\arraystretch}{0.95}
    \resizebox{\textwidth}{!}{%
      \begin{tabular}{lccccc|ccccc|ccccc}
        \toprule
        & \multicolumn{5}{c}{HM-ReasoningBench}
        & \multicolumn{5}{c}{AIME25}
        & \multicolumn{5}{c}{Apex} \\
        \cmidrule(lr){2-6}
        \cmidrule(lr){7-11}
        \cmidrule(lr){12-16}
        Model
        & DS & ICL & H & M & SSR
        & DS & ICL & H & M & SSR
        & DS & ICL & H & M & SSR \\
        \midrule
        Qwen3-8B
        & 63.80 & 64.20 & 66.00 & 65.40 & \textbf{68.60}
        & 67.33 & 68.00 & 70.67 & 69.33 & \textbf{74.00}
        & 8.16  & 8.16  & 8.57  & 8.98  & \textbf{13.06} \\
        Qwen3-14B
        & 67.40 & 67.60 & 69.40 & 68.20 & \textbf{70.20}
        & 70.66 & 70.33 & 72.67 & 72.00 & \textbf{74.67}
        & 11.02 & 11.43 & 13.47 & 12.65 & \textbf{14.69} \\
        R1-Distill-7B
        & 49.20 & 49.80 & 51.00 & 50.80 & \textbf{52.40}
        & 42.00 & 48.00 & 51.33 & 52.00 & \textbf{53.13}
        & 4.08  & 4.90  & 6.53  & 6.12  & \textbf{7.75} \\
        \bottomrule
      \end{tabular}
    }%
    \endgroup
  \end{center}
  \vskip -0.1in
\end{table*}

\begin{table*}[t]
  \caption{
    Ablation study of retrieval routes in Selective Strategy Retrieval (SSR)
    across three benchmarks.
    Best result is shown in bold.
  }
  \label{tab:ablation}
  \begin{center}
    \begingroup
    \scriptsize
    \setlength{\tabcolsep}{3pt}
    \renewcommand{\arraystretch}{0.95}
    \resizebox{\textwidth}{!}{%
      \begin{tabular}{lcccc|cccc|cccc}
        \toprule
        & \multicolumn{4}{c}{HM-ReasoningBench}
        & \multicolumn{4}{c}{AIME25}
        & \multicolumn{4}{c}{Apex} \\
        \cmidrule(lr){2-5}
        \cmidrule(lr){6-9}
        \cmidrule(lr){10-13}
        Model
        & SSR & w/o Cat & w/o Tran & w/o Sem
        & SSR & w/o Cat & w/o Tran & w/o Sem
        & SSR & w/o Cat & w/o Tran & w/o Sem \\
        \midrule
        Qwen3-8B
        & \textbf{68.60} & 66.00 & 63.60 & 67.20
        & \textbf{74.00} & 70.67 & 67.33 & 72.67
        & \textbf{13.06} & 12.24 & 11.02 & 13.06 \\
        Qwen3-14B
        & \textbf{70.20} & 67.60 & 65.00 & 69.00
        & \textbf{74.67} & 71.33 & 68.67 & 73.33
        & \textbf{14.69} & 13.06 & 11.84 & 13.88 \\
        R1-Distill-7B
        & \textbf{52.40} & 50.60 & 48.80 & 51.20
        & \textbf{53.13} & 52.00 & 50.67 & 51.33
        & \textbf{7.75}  & 6.94  & 6.12  & 7.35 \\
        \bottomrule
      \end{tabular}
    }%
    \endgroup
  \end{center}
  \vskip -0.1in
\end{table*}

\section{Experiments}
\label{sec:experiments}

We evaluate whether executability-aware strategy selection, implemented by Selective
Strategy Retrieval (SSR), reliably improves mathematical reasoning.
Our experiments address three questions:
(i) does SSR consistently improve accuracy across datasets and models,
(ii) are its key components necessary, and
(iii) when and why human-derived strategies are most effective.

\subsection{Experimental Setup}
\label{subsec:setup}

\noindent\textbf{Datasets.}
We evaluate on three benchmarks spanning paired analysis, competition-style reasoning,
and extreme difficulty.
\textbf{HM-ReasoningBench} is used to construct the strategy knowledge graph and is
evaluated on a held-out test split.
\textbf{AIME 2025}~\cite{maa_aime_2025_misc} contains competition-level problems requiring
multi-step symbolic reasoning.
\textbf{MathArena Apex}~\cite{balunovic_srimatharena_2025} consists of highly challenging
final-answer problems on which even strong models have low success rates, serving as a
stress test for compositional reasoning.

\noindent\textbf{Models.}
We evaluate \textbf{Qwen3-8B}, \textbf{Qwen3-14B}, and \textbf{DeepSeek-R1-Distill-Qwen-7B}.
All models use the same configuration (max context 32{,}768; temperature 0.7).

\noindent\textbf{Metric and verification.}
We report exact-match accuracy.
For proof-oriented problems, we use GPT-5.1 to verify mathematical equivalence between
model outputs and references; the verification prompt is provided in
Appendix~\ref{app:av_prompt}.

\noindent\textbf{Reproducibility.}
All reported results are averaged over 5 independent runs with different random seeds.
We report mean accuracy throughout.

\subsection{Baselines}
\label{subsec:baselines}

All methods share the same prompting format and differ only in how guidance is sourced.
\textbf{Direct Solving (DS)} solves the problem without external guidance;
\textbf{In-Context Learning (ICL)} provides one worked example (more examples did not help
and sometimes degraded performance);
\textbf{Human-Only Guidance (H)} uses strategy hints extracted from human solutions; and
\textbf{Model-Only Guidance (M)} uses strategy hints extracted from model solutions.
In our implementation, the worked example used by ICL is itself selected by semantic
retrieval; additional retrieval-based comparisons are reported in
Appendix~\ref{app:retrieval_baselines}.

We also compare against stronger inference-time baselines, including
Self-Consistency (SC), Least-to-Most Prompting (L2M), and Tree-of-Thoughts (ToT),
which allocate additional test-time computation through sampling or search.
\subsection{Main Results}
\label{subsec:main_results}

Table~\ref{tab:main_results} reports accuracy across datasets.
Three consistent patterns emerge.
First, strategy guidance improves over DS in all settings, indicating that abstract
strategy hints are generally usable by compact reasoning models.
Second, \textbf{SSR consistently achieves the best performance}, outperforming both
single-source guidance (H/M) and ICL, which rules out gains from merely adding more context.
Third, SSR’s relative advantage increases with benchmark difficulty, consistent with our
executability analysis. Comparisons with stronger inference-time baselines, including self-consistency,
least-to-most prompting, and Tree-of-Thoughts, are reported in
Appendix~\ref{app:additional_baselines}. \textbf{Notably, SSR achieves these gains using a single guided generation per problem,
whereas these baselines allocate substantially more test-time computation.}

On \textsc{HM-ReasoningBench} and \textsc{AIME25}, both H and M improve over DS, while SSR
yields further gains (e.g., Qwen3-8B: 63.80 $\rightarrow$ 66.00/65.40 $\rightarrow$ 68.60 on
the former, and 67.33 $\rightarrow$ 70.67/69.33 $\rightarrow$ 74.00 on the latter).
On the hardest benchmark \textsc{Apex}, SSR’s gains are largest (e.g., Qwen3-8B:
8.16 $\rightarrow$ 13.06), reflecting the amplified impact of executability mismatches in
long-horizon problems.

Across datasets, H slightly outperforms M on average.
\textbf{SSR consistently improves over both} by selecting and combining strategies in a
context- and source-aware manner.
Qualitative examples illustrating how SSR yields more coherent reasoning trajectories are
provided in Appendix~\ref{app:qual_examples}.

To further isolate whether the gains come from executability-aware selection rather than
retrieval alone, Table~\ref{tab:retrieval_main} compares SSR against end-to-end BM25,
dense retrieval, and MMR-style semantic reranking baselines under the same prompt budget
and decoding protocol.

\begin{table}[t]
  \caption{End-to-end retrieval baselines on AIME25 (accuracy \%).}
  \label{tab:retrieval_main}
  \begin{center}
    \scriptsize
    \begin{tabular}{lccc}
      \toprule
      Method & Qwen3-8B & Qwen3-14B & R1-Distill-7B \\
      \midrule
      BM25 retrieval & 67.33 & 69.00 & 47.33 \\
      Dense retrieval & 68.00 & 70.33 & 48.00 \\
      MMR semantic reranking & 68.67 & 71.00 & 49.33 \\
      \textbf{SSR} & \textbf{74.00} & \textbf{74.67} & \textbf{53.13} \\
      \bottomrule
    \end{tabular}
  \end{center}
  \vskip -0.15in
\end{table}

SSR remains best across all three models, showing that the gains are not explained by
semantic retrieval quality alone.

\subsection{Ablation: Is Multi-Route Retrieval Necessary?}
\label{subsec:ablation}

SSR constructs candidates via three retrieval routes corresponding to distinct
executability signals.
We ablate each route in turn while keeping all other components fixed:
Route A (category-conditioned), Route B (problem-transfer), and Route C (semantic fallback).

As shown in Table~\ref{tab:ablation}, removing any route consistently degrades performance,
indicating that no single signal is sufficient.
Removing Route B causes the largest drop (e.g., Qwen3-14B on \textsc{Apex}:
14.69 $\rightarrow$ 11.84), highlighting the importance of fine-grained,
context-dependent transfer.
Removing Route A also leads to clear declines (e.g., Qwen3-8B on \textsc{AIME25}:
74.00 $\rightarrow$ 70.67), while removing Route C yields smaller but consistent drops,
reflecting its role in maintaining coverage when executability evidence is sparse.

\subsection{Analysis: When Does Human Guidance Help Most?}
\label{subsec:human_analysis}

Our strategy-level diagnosis predicts that human-derived guidance helps most when failures
are driven by missing \emph{global structure} (e.g., absent decomposition, constraints, or
case splits) rather than local symbolic slips.
We test this prediction using topic-level and failure-mode analyses.

\begin{figure}[h]
  \begin{center}
    \centerline{\includegraphics[width=0.85\linewidth]{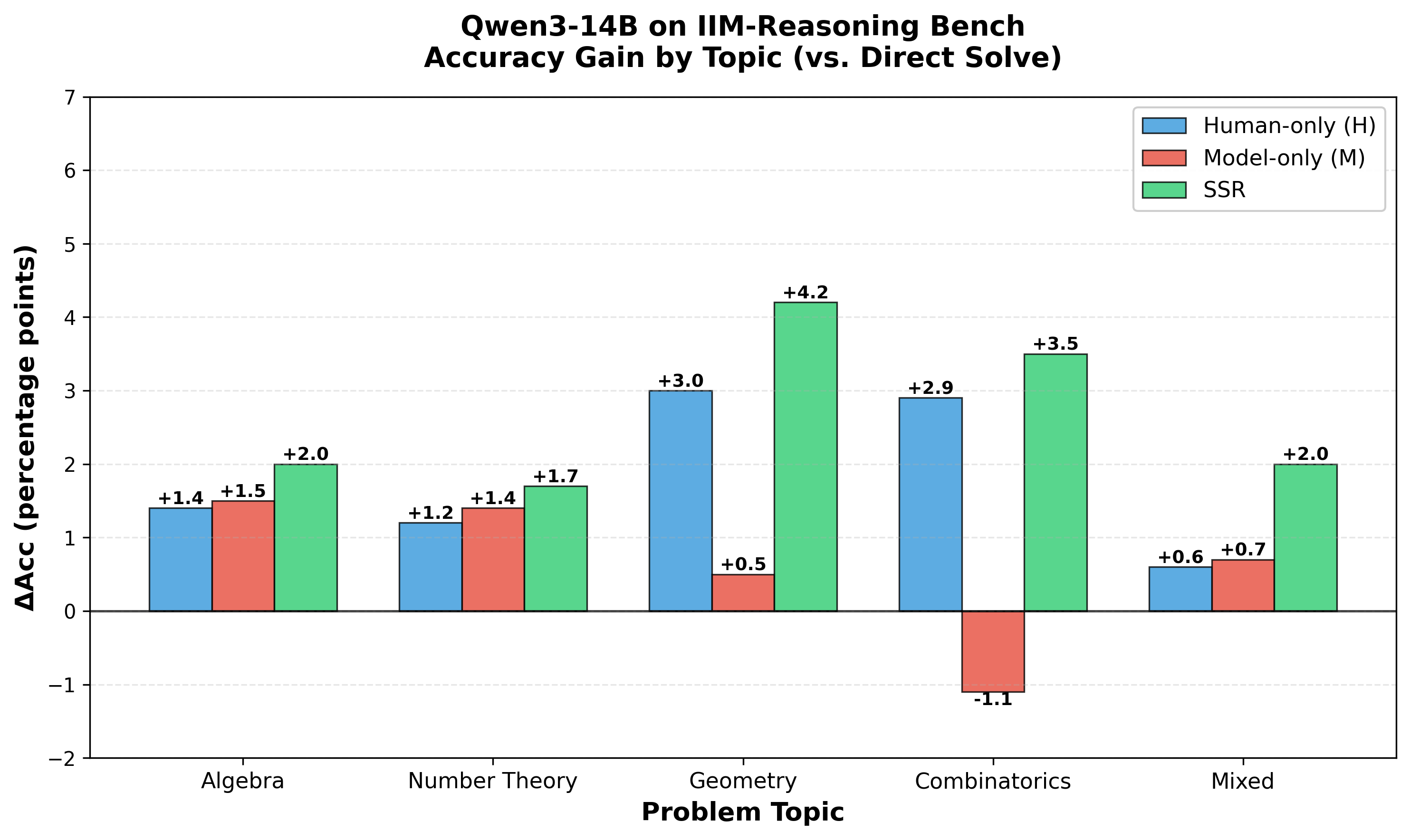}}
    \caption{
      Topic-wise gains on \textsc{HM-ReasoningBench} using
      Qwen3-14B. Results exhibit domain-dependent behavior.
    }
    \label{fig:topic_gain}
  \end{center}
\end{figure}

\noindent\textbf{Topic-level.}
Figure~\ref{fig:topic_gain} reports gains over DS on \textsc{HM-ReasoningBench}
(Qwen3-14B).
Human guidance yields the largest gains in geometry and combinatorics, while model guidance
is weaker and can degrade performance.
In algebra and number theory, source effects are smaller and both H and M provide modest
gains.
Across topics, \textbf{SSR matches or exceeds the stronger source}, confirming that
source effectiveness is context-dependent.

This pattern can be understood through two complementary task properties.
When the main bottleneck is finding the right \emph{high-level reformulation},
human-derived structural guidance is often more executable.
When success instead depends on carrying out a long sequence of explicit intermediate
operations, model-derived procedural guidance becomes more reliable.
Appendix~\ref{app:style_pool} reports a controlled style-pool analysis supporting this
task-conditional interpretation.

\noindent\textbf{Failure modes.}
We further analyze failure modes of incorrect solutions by categorizing them into
\emph{structural reasoning failures} and \emph{algebraic manipulation errors}.
Human guidance predominantly reduces structural failures, while model guidance is more
effective at mitigating algebraic errors.
SSR reduces both failure types, consistent with executability-aware selection that
combines complementary structural and procedural signals (see Appendix~\ref{app:failure_mode}
for definitions and quantitative breakdowns).

\subsection{Efficiency and Context Budget}
\label{subsec:efficiency}

\begin{figure}[h]
  \begin{center}
    \centerline{\includegraphics[width=0.85\linewidth]{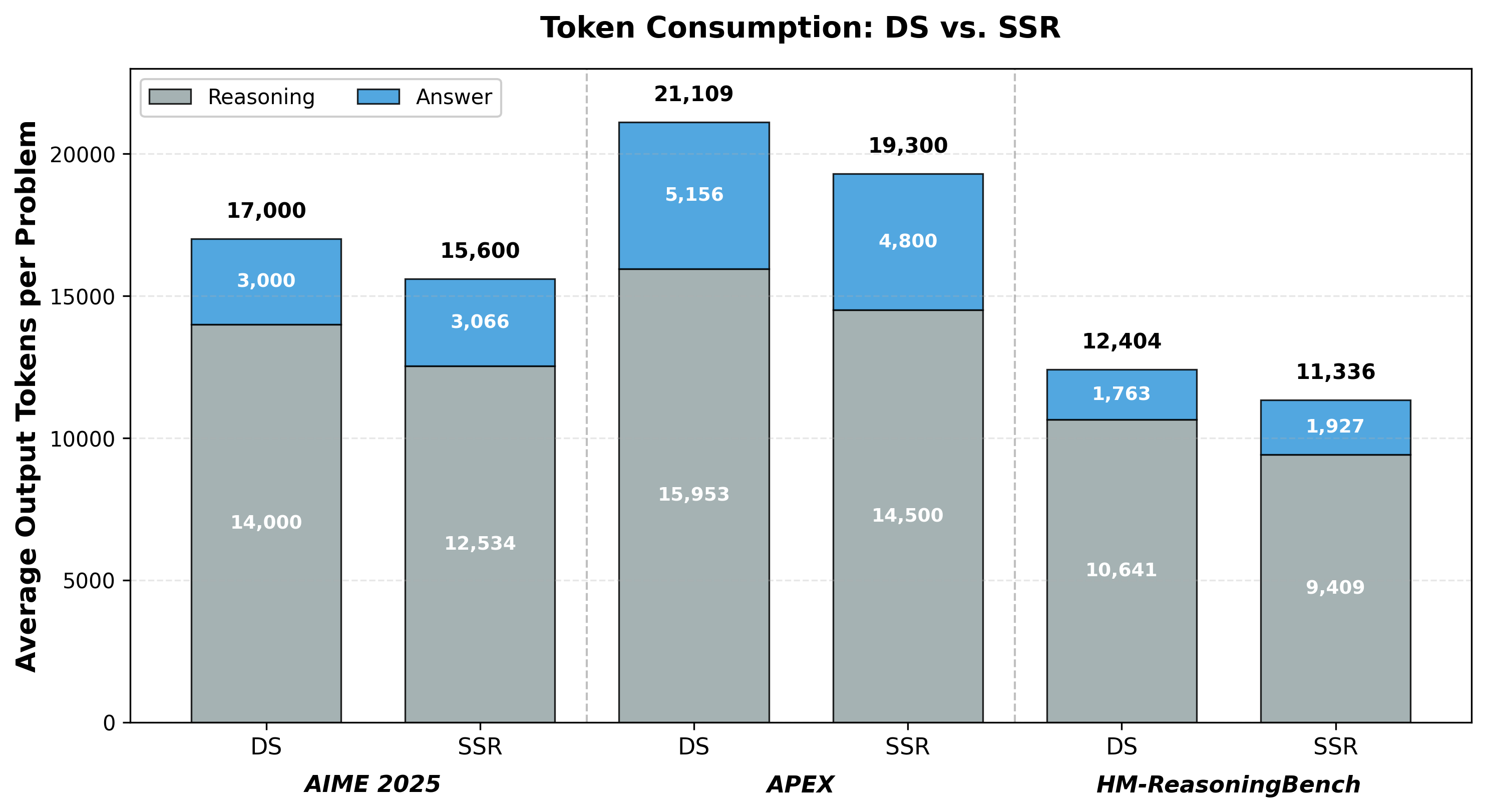}}
    \caption{
      Average output token consumption per problem under direct solving (DS)
      and Selective Strategy Retrieval (SSR) using Qwen3-14B, decomposed into
      reasoning and final-answer tokens.
    }
    \label{fig:token_consumption}
  \end{center}
\end{figure}

We measure output token consumption under DS and SSR using Qwen3-14B across all three
benchmarks.
Figure~\ref{fig:token_consumption} shows that SSR reduces total output tokens relative to
DS, with reductions concentrated in reasoning tokens.
This suggests that SSR lowers \emph{online generation burden} by steering models away from
unproductive exploration rather than eliciting longer reasoning traces.
Reductions are largest on \textsc{Apex} and \textsc{HM-ReasoningBench}, which require
longer-horizon reasoning; on \textsc{AIME25} they are smaller but consistent.
This figure should therefore be interpreted as an online token comparison rather than a
full end-to-end cost analysis; Appendix~\ref{app:cost_breakdown} reports the additional
offline and online retrieval costs, together with a calibrated comparison to ToT.
In particular, SSR trades a modest one-time offline build cost for low online overhead,
whereas ToT is training-free but substantially more expensive at test time.

\paragraph{Strategy adherence.}
\label{subsec:adherence}
To verify that SSR’s gains reflect meaningful strategy execution rather than prompt length,
we conduct an adherence-style sanity check.
Correct solutions are substantially more likely to correctly instantiate at least one
provided strategy, supporting the interpretation that SSR improves executability rather
than verbosity.
Full protocol and results are provided in Appendix~\ref{app:adherence}.





\section{Conclusion}

We revisited example-based guidance for mathematical reasoning from a strategy-level
perspective and identified a systematic gap between \emph{strategy usage} and
\emph{strategy executability}: strategies that commonly appear in correct solutions are
not necessarily those a target model can reliably execute as guidance, explaining the
instability of guidance and the limits of uniform imitation.

To address this failure mode, we proposed \textbf{Selective Strategy Retrieval (SSR)}, a
test-time framework that prioritizes strategies with stronger empirical evidence of
executability and consistently outperforms direct solving, in-context learning, and
single-source strategy guidance across multiple benchmarks and compact reasoning models.

Additional analyses further sharpen this picture.
They show that the usage--executability gap persists on stronger GPT-series models,
that executability estimates remain stable under higher-trial evaluation, and that SSR's
advantages cannot be reduced to semantic retrieval alone.
They also clarify the deployment trade-off of SSR: the method incurs a nonzero but
lightweight offline build cost, while keeping online reasoning substantially cheaper than
search-based alternatives such as ToT.

Although we study mathematics as a controlled testbed with paired human/model solutions
and reliable correctness checking, the underlying distinction between usage and
executability is more general.
Whenever external guidance must be translated into executable intermediate reasoning by a
target model, semantically appropriate strategies may still fail if they are not
operationally aligned with that model.
Extending this perspective to broader scientific, coding, and logical reasoning domains
is a natural next step.

More broadly, our findings suggest that reasoning guidance should be evaluated
\emph{model-relatively}: the usefulness of a strategy depends on whether the target model
can operationalize it under the given context, motivating executability-aware evaluation
and guidance mechanisms grounded in context-dependent effectiveness.

\newpage

\section*{Impact Statement}

This paper presents work whose goal is to advance the understanding of how
reasoning guidance interacts with model behavior in mathematical problem solving.
We introduce the notion of \emph{strategy executability} to distinguish between
reasoning strategies that appear in successful solutions and those that a target
model can reliably operationalize when provided as guidance under fixed inference
conditions.
Building on this perspective, we develop Selective Strategy Retrieval (SSR), a
lightweight inference-time framework that improves robustness by selecting and
combining strategies based on empirical executability signals rather than surface
correctness or prevalence alone.

Beyond its immediate implications for inference-time guidance, this work has
potential relevance for future research on model training and evaluation.
Our analysis reveals a systematic and structured mismatch between human-written
and model-generated solutions: although both may reach correct answers, they
exhibit consistent differences in the types of reasoning strategies they employ
and successfully execute.
This dissociation suggests that current training data distributions and learning
objectives may implicitly reinforce certain procedural or algebraic reasoning
patterns while under-representing more abstract, structural, or conceptually
driven strategies commonly used by humans.
From this perspective, human--model disagreement is not merely an inference-time
artifact, but a diagnostic signal of imbalance in how reasoning behaviors are
learned and reinforced during training.

While this paper does not propose changes to model architectures, training
procedures, or supervision schemes, the concept of strategy executability may
inform future efforts to design training curricula, auxiliary objectives, or
evaluation protocols that better reflect whether models can reliably execute
different classes of reasoning strategies under controlled conditions.
More broadly, our findings highlight the importance of evaluating reasoning
capabilities not only by final-answer correctness, but by the operational
usability of intermediate reasoning abstractions.

The expected societal impact of this work is indirect.
By clarifying when and why reasoning guidance succeeds or fails, our contributions
may support the development of more reliable, interpretable, and controllable
reasoning systems for educational, scientific, and analytical applications.
This work does not involve human subjects, personal data, or real-world
decision-making systems, and it does not introduce new risks beyond those commonly
associated with foundational research in machine learning.

\bibliography{reference}

@article{wei2022chain,
  title={Chain-of-thought prompting elicits reasoning in large language models},
  author={Wei, Jason and Wang, Xuezhi and Schuurmans, Dale and Bosma, Maarten and Xia, Fei and Chi, Ed and Le, Quoc V and Zhou, Denny and others},
  journal={Advances in neural information processing systems},
  volume={35},
  pages={24824--24837},
  year={2022}
}

@article{brown2020language,
  title={Language models are few-shot learners},
  author={Brown, Tom and Mann, Benjamin and Ryder, Nick and Subbiah, Melanie and Kaplan, Jared D and Dhariwal, Prafulla and Neelakantan, Arvind and Shyam, Pranav and Sastry, Girish and Askell, Amanda and others},
  journal={Advances in neural information processing systems},
  volume={33},
  pages={1877--1901},
  year={2020}
}

@inproceedings{yao2022react,
  title={React: Synergizing reasoning and acting in language models},
  author={Yao, Shunyu and Zhao, Jeffrey and Yu, Dian and Du, Nan and Shafran, Izhak and Narasimhan, Karthik R and Cao, Yuan},
  booktitle={The eleventh international conference on learning representations},
  year={2022}
}

@article{madsen2024self,
  title={Are self-explanations from Large Language Models faithful?},
  author={Madsen, Andreas and Chandar, Sarath and Reddy, Siva},
  journal={arXiv preprint arXiv:2401.07927},
  year={2024}
}

@article{guo2025deepseek,
  title={Deepseek-r1: Incentivizing reasoning capability in llms via reinforcement learning},
  author={Guo, Daya and Yang, Dejian and Zhang, Haowei and Song, Junxiao and Zhang, Ruoyu and Xu, Runxin and Zhu, Qihao and Ma, Shirong and Wang, Peiyi and Bi, Xiao and others},
  journal={arXiv preprint arXiv:2501.12948},
  year={2025}
}

@article{lewkowycz2022solving,
  title={Solving quantitative reasoning problems with language models},
  author={Lewkowycz, Aitor and Andreassen, Anders and Dohan, David and Dyer, Ethan and Michalewski, Henryk and Ramasesh, Vinay and Slone, Ambrose and Anil, Cem and Schlag, Imanol and Gutman-Solo, Theo and others},
  journal={Advances in neural information processing systems},
  volume={35},
  pages={3843--3857},
  year={2022}
}

@article{kojima2022large,
  title={Large language models are zero-shot reasoners},
  author={Kojima, Takeshi and Gu, Shixiang Shane and Reid, Machel and Matsuo, Yutaka and Iwasawa, Yusuke},
  journal={Advances in neural information processing systems},
  volume={35},
  pages={22199--22213},
  year={2022}
}

@article{achiam2023gpt,
  title={Gpt-4 technical report},
  author={Achiam, Josh and Adler, Steven and Agarwal, Sandhini and Ahmad, Lama and Akkaya, Ilge and Aleman, Florencia Leoni and Almeida, Diogo and Altenschmidt, Janko and Altman, Sam and Anadkat, Shyamal and others},
  journal={arXiv preprint arXiv:2303.08774},
  year={2023}
}

@article{creswell2022faithful,
  title={Faithful reasoning using large language models},
  author={Creswell, Antonia and Shanahan, Murray},
  journal={arXiv preprint arXiv:2208.14271},
  year={2022}
}

@article{zelikman2022star,
  title={Star: Bootstrapping reasoning with reasoning},
  author={Zelikman, Eric and Wu, Yuhuai and Mu, Jesse and Goodman, Noah},
  journal={Advances in Neural Information Processing Systems},
  volume={35},
  pages={15476--15488},
  year={2022}
}

@article{yao2023tree,
  title={Tree of thoughts: Deliberate problem solving with large language models},
  author={Yao, Shunyu and Yu, Dian and Zhao, Jeffrey and Shafran, Izhak and Griffiths, Tom and Cao, Yuan and Narasimhan, Karthik},
  journal={Advances in neural information processing systems},
  volume={36},
  pages={11809--11822},
  year={2023}
}

@article{madaan2023self,
  title={Self-refine: Iterative refinement with self-feedback},
  author={Madaan, Aman and Tandon, Niket and Gupta, Prakhar and Hallinan, Skyler and Gao, Luyu and Wiegreffe, Sarah and Alon, Uri and Dziri, Nouha and Prabhumoye, Shrimai and Yang, Yiming and others},
  journal={Advances in Neural Information Processing Systems},
  volume={36},
  pages={46534--46594},
  year={2023}
}

@article{mukherjee2025premise,
  title={Premise-augmented reasoning chains improve error identification in math reasoning with llms},
  author={Mukherjee, Sagnik and Chinta, Abhinav and Kim, Takyoung and Sharma, Tarun Anoop and Hakkani-T{\"u}r, Dilek},
  journal={arXiv preprint arXiv:2502.02362},
  year={2025}
}

@article{chowdhury2025zero,
  title={Zero-shot verification-guided chain of thoughts},
  author={Chowdhury, Jishnu Ray and Caragea, Cornelia},
  journal={arXiv preprint arXiv:2501.13122},
  year={2025}
}

@inproceedings{diao2024active,
  title={Active prompting with chain-of-thought for large language models},
  author={Diao, Shizhe and Wang, Pengcheng and Lin, Yong and Pan, Rui and Liu, Xiang and Zhang, Tong},
  booktitle={Proceedings of the 62nd Annual Meeting of the Association for Computational Linguistics (Volume 1: Long Papers)},
  pages={1330--1350},
  year={2024}
}

@article{fernando2023promptbreeder,
  title={Promptbreeder: Self-referential self-improvement via prompt evolution},
  author={Fernando, Chrisantha and Banarse, Dylan and Michalewski, Henryk and Osindero, Simon and Rockt{\"a}schel, Tim},
  journal={arXiv preprint arXiv:2309.16797},
  year={2023}
}

@article{wang2022self,
  title={Self-consistency improves chain of thought reasoning in language models},
  author={Wang, Xuezhi and Wei, Jason and Schuurmans, Dale and Le, Quoc and Chi, Ed and Narang, Sharan and Chowdhery, Aakanksha and Zhou, Denny},
  journal={arXiv preprint arXiv:2203.11171},
  year={2022}
}

@inproceedings{xu2024pride,
  title={Pride and prejudice: LLM amplifies self-bias in self-refinement},
  author={Xu, Wenda and Zhu, Guanglei and Zhao, Xuandong and Pan, Liangming and Li, Lei and Wang, William},
  booktitle={Proceedings of the 62nd Annual Meeting of the Association for Computational Linguistics (Volume 1: Long Papers)},
  pages={15474--15492},
  year={2024}
}

@article{wu2025more,
  title={When more is less: Understanding chain-of-thought length in llms},
  author={Wu, Yuyang and Wang, Yifei and Ye, Ziyu and Du, Tianqi and Jegelka, Stefanie and Wang, Yisen},
  journal={arXiv preprint arXiv:2502.07266},
  year={2025}
}

@article{gao2024omni,
  title={Omni-math: A universal olympiad level mathematic benchmark for large language models},
  author={Gao, Bofei and Song, Feifan and Yang, Zhe and Cai, Zefan and Miao, Yibo and Dong, Qingxiu and Li, Lei and Ma, Chenghao and Chen, Liang and Xu, Runxin and others},
  journal={arXiv preprint arXiv:2410.07985},
  year={2024}
}

@article{yue2024harp,
  title={HARP: A challenging human-annotated math reasoning benchmark},
  author={Yue, Albert S and Madaan, Lovish and Moskovitz, Ted and Strouse, DJ and Singh, Aaditya K},
  journal={arXiv preprint arXiv:2412.08819},
  year={2024}
}

@inproceedings{liu2022generated,
  title={Generated knowledge prompting for commonsense reasoning},
  author={Liu, Jiacheng and Liu, Alisa and Lu, Ximing and Welleck, Sean and West, Peter and Le Bras, Ronan and Choi, Yejin and Hajishirzi, Hannaneh},
  booktitle={Proceedings of the 60th annual meeting of the association for computational linguistics (volume 1: long papers)},
  pages={3154--3169},
  year={2022}
}

@article{zhou2022least,
  title={Least-to-most prompting enables complex reasoning in large language models},
  author={Zhou, Denny and Sch{\"a}rli, Nathanael and Hou, Le and Wei, Jason and Scales, Nathan and Wang, Xuezhi and Schuurmans, Dale and Cui, Claire and Bousquet, Olivier and Le, Quoc and others},
  journal={arXiv preprint arXiv:2205.10625},
  year={2022}
}

@article{shum2023automatic,
  title={Automatic prompt augmentation and selection with chain-of-thought from labeled data},
  author={Shum, KaShun and Diao, Shizhe and Zhang, Tong},
  journal={arXiv preprint arXiv:2302.12822},
  year={2023}
}

@inproceedings{rubin2022learning,
  title={Learning to retrieve prompts for in-context learning},
  author={Rubin, Ohad and Herzig, Jonathan and Berant, Jonathan},
  booktitle={Proceedings of the 2022 conference of the North American chapter of the association for computational linguistics: human language technologies},
  pages={2655--2671},
  year={2022}
}

@inproceedings{wu2023self,
  title={Self-adaptive in-context learning: An information compression perspective for in-context example selection and ordering},
  author={Wu, Zhiyong and Wang, Yaoxiang and Ye, Jiacheng and Kong, Lingpeng},
  booktitle={Proceedings of the 61st Annual Meeting of the Association for Computational Linguistics (Volume 1: Long Papers)},
  pages={1423--1436},
  year={2023}
}

@article{cobbe2021training,
  title={Training verifiers to solve math word problems},
  author={Cobbe, Karl and Kosaraju, Vineet and Bavarian, Mohammad and Chen, Mark and Jun, Heewoo and Kaiser, Lukasz and Plappert, Matthias and Tworek, Jerry and Hilton, Jacob and Nakano, Reiichiro and others},
  journal={arXiv preprint arXiv:2110.14168},
  year={2021}
}

@article{zhang2025booststep,
  title={Booststep: Boosting mathematical capability of large language models via improved single-step reasoning},
  author={Zhang, Beichen and Liu, Yuhong and Dong, Xiaoyi and Zang, Yuhang and Zhang, Pan and Duan, Haodong and Cao, Yuhang and Lin, Dahua and Wang, Jiaqi},
  journal={arXiv preprint arXiv:2501.03226},
  year={2025}
}

@inproceedings{cao2025step,
  title = {Step Guided Reasoning: Improving Mathematical Reasoning using Guidance Generation and Step Reasoning},
  author = {Cao, Lang and Zou, Yingtian and Peng, Chao and Chen, Renhong and Ning, Wu and Li, Yitong},
  booktitle = {Proceedings of the 2025 Conference on Empirical Methods in Natural Language Processing},
  pages = {21101--21118},
  year = {2025},
  publisher = {Association for Computational Linguistics},
  doi = {10.18653/v1/2025.emnlp-main.1068}
}

@inproceedings{jiang2025makes,
  title={What makes a good reasoning chain? uncovering structural patterns in long chain-of-thought reasoning},
  author={Jiang, Gangwei and Liu, Yahui and Li, Zhaoyi and Bi, Wei and Zhang, Fuzheng and Song, Linqi and Wei, Ying and Lian, Defu},
  booktitle={Proceedings of the 2025 Conference on Empirical Methods in Natural Language Processing},
  pages={6501--6525},
  year={2025}
}

@article{munkhbat2025self,
  title={Self-training elicits concise reasoning in large language models},
  author={Munkhbat, Tergel and Ho, Namgyu and Kim, Seo Hyun and Yang, Yongjin and Kim, Yujin and Yun, Se-Young},
  journal={arXiv preprint arXiv:2502.20122},
  year={2025}
}

@inproceedings{yu2025chain,
  title={Chain-of-reasoning: Towards unified mathematical reasoning in large language models via a multi-paradigm perspective},
  author={Yu, Yiyao and Zhang, Yuxiang and Zhang, Dongdong and Liang, Xiao and Zhang, Hengyuan and Zhang, Xingxing and Khademi, Mahmoud and Awadalla, Hany Hassan and Wang, Junjie and Yang, Yujiu and others},
  booktitle={Proceedings of the 63rd Annual Meeting of the Association for Computational Linguistics (Volume 1: Long Papers)},
  pages={24914--24937},
  year={2025}
}

@misc{maa_aime_2025_misc,
  author       = {{Mathematical Association of America}},
  title        = {American Invitational Mathematics Examination (AIME)},
  howpublished = {\url{https://maa.org/maa-invitational-competitions/}},
  year         = {2025},
  note         = {Accessed: 2025-08-19}
}

@misc{balunovic_srimatharena_2025,
  title = {MathArena: Evaluating LLMs on Uncontaminated Math Competitions},
  author = {Mislav Balunović and Jasper Dekoninck and Ivo Petrov and Nikola Jovanović and Martin Vechev},
  copyright = {MIT},
  url = {https://matharena.ai/},
  publisher = {SRI Lab, ETH Zurich},
  month = feb,
  year = {2025},
}

@book{zeitz2016art,
  title={The art and craft of problem solving},
  author={Zeitz, Paul},
  year={2016},
  publisher={John Wiley \& Sons}
}

@book{engel1998problem,
  title={Problem-solving strategies},
  author={Engel, Arthur},
  year={1998},
  publisher={Springer}
}

@article{polya1957solve,
  title={How to solve it},
  author={Polya, George},
  year={1957},
  publisher={Princeton Press}
}

@article{xu2025teaching,
  title={Teaching llms according to their aptitude: Adaptive reasoning for mathematical problem solving},
  author={Xu, Xin and Xu, Yan and Chen, Tianhao and Yan, Yuchen and Liu, Chengwu and Chen, Zaoyu and Wang, Yufei and Yin, Yichun and Wang, Yasheng and Shang, Lifeng and others},
  journal={arXiv preprint arXiv:2502.12022},
  year={2025}
}

@article{qi2025plan,
  title={Plan before Solving: Problem-Aware Strategy Routing for Mathematical Reasoning with LLMs},
  author={Qi, Shihao and Ma, Jie and Yin, Ziang and Zhang, Lingling and Zhang, Jian and Liu, Jun and Tian, Feng and Liu, Tongliang},
  journal={arXiv preprint arXiv:2509.24377},
  year={2025}
}

@article{simon1971human,
  title={Human problem solving: The state of the theory in 1970.},
  author={Simon, Herbert A and Newell, Allen},
  journal={American psychologist},
  volume={26},
  number={2},
  pages={145},
  year={1971},
  publisher={American Psychological Association}
}

@article{chi2006laboratory,
  title={Laboratory methods for assessing experts’ and novices’ knowledge},
  author={Chi, Michelene TH},
  journal={The Cambridge handbook of expertise and expert performance},
  pages={167--184},
  year={2006}
}

@article{larkin1980expert,
  title={Expert and novice performance in solving physics problems},
  author={Larkin, Jill and McDermott, John and Simon, Dorothea P and Simon, Herbert A},
  journal={Science},
  volume={208},
  number={4450},
  pages={1335--1342},
  year={1980},
  publisher={American Association for the Advancement of Science}
}

@article{mahdavi2025brains,
  title={Brains vs. bytes: Evaluating llm proficiency in olympiad mathematics},
  author={Mahdavi, Hamed and Hashemi, Alireza and Daliri, Majid and Mohammadipour, Pegah and Farhadi, Alireza and Malek, Samira and Yazdanifard, Yekta and Khasahmadi, Amir and Honavar, Vasant},
  journal={arXiv preprint arXiv:2504.01995},
  year={2025}
}

@article{trinh2024solving,
  title={Solving olympiad geometry without human demonstrations},
  author={Trinh, Trieu H and Wu, Yuhuai and Le, Quoc V and He, He and Luong, Thang},
  journal={Nature},
  volume={625},
  number={7995},
  pages={476--482},
  year={2024},
  publisher={Nature Publishing Group UK London}
}

@article{ruis2411procedural,
  title={Procedural Knowledge in Pretraining Drives Reasoning in Large Language Models. arXiv 2024},
  author={Ruis, Laura and Mozes, Maximilian and Bae, Juhan and Kamalakara, Siddhartha Rao and Talupuru, Dwarak and Locatelli, Acyr and Kirk, Robert and Rockt{\"a}schel, Tim and Grefenstette, Edward and Bartolo, Max},
  journal={arXiv preprint arXiv:2411.12580}
}

@article{hu2025coarse,
  title={Coarse-to-fine process reward modeling for mathematical reasoning},
  author={Hu, Yulan and Ouyang, Sheng and Zhao, Jinman and Liu, Yong},
  journal={arXiv preprint arXiv:2501.13622},
  year={2025}
}

@article{younsi2025accurate,
  title={Accurate and diverse llm mathematical reasoning via automated prm-guided gflownets},
  author={Younsi, Adam and Attia, Ahmed and Abubaker, Abdalgader and Seddik, Mohamed El Amine and Hacid, Hakim and Lahlou, Salem},
  journal={arXiv preprint arXiv:2504.19981},
  year={2025}
}
\bibliographystyle{icml2026}

\newpage

\appendix
\onecolumn
\section{Implementation Details}

\subsection{More Details for HM-ReasoningBench}
\label{app:dataset}

\noindent\textbf{Dataset Overview.}
HM-ReasoningBench is a large-scale mathematical reasoning benchmark constructed from two complementary sources: \emph{OmniMATH} and \emph{HARP}.
After removing exact duplicate problems by text, the final benchmark contains \textbf{4,895 unique problems}.
Among them, \textbf{310 problems} are drawn from HARP, while the remaining \textbf{4,585 problems} are sourced from OmniMATH.

\noindent\textbf{Difficulty Annotation (Level).}
Each problem is assigned a discrete difficulty level ranging from Level~1 (easiest) to Level~9 (hardest).
In practice, we observe substantial variation in problem difficulty even within the same competition
or source, making it insufficient to rely on competition-level tiers or inherited difficulty labels.
To obtain a more objective, instance-level assessment, we perform a unified re-annotation of problem
difficulty across all sources.

Concretely, difficulty is assigned under a shared reference framework that anchors problems to a
common difficulty scale spanning typical Olympiad-style reasoning tasks.
GPT-5.1 is used as a calibrated assessor to map individual problems onto this scale, guided by
cross-competition comparisons rather than source-specific context.
This procedure enforces consistency across heterogeneous sources and enables meaningful
cross-source difficulty analysis.
As a result, the difficulty distribution concentrates in mid-to-high ranges, with Level~5--7
accounting for the majority of problems.

\begin{table}[b]
  \caption{
    Difficulty level distribution in HM-ReasoningBench after GPT-5.1 re-annotation.
  }
  \label{tab:difficulty_dist}
  \begin{center}
    \begin{small}
      \begin{sc}
        \begin{tabular}{lrr}
          \toprule
          Difficulty Level & Count & Percentage \\
          \midrule
          Level 1 & 78   & 1.6\% \\
          Level 2 & 303  & 6.1\% \\
          Level 3 & 386  & 7.8\% \\
          Level 4 & 742  & 15.0\% \\
          Level 5 & 1140 & 23.1\% \\
          Level 6 & 1197 & 24.3\% \\
          Level 7 & 908  & 18.4\% \\
          Level 8 & 177  & 3.6\% \\
          Level 9 & 1    & 0.0\% \\
          \bottomrule
        \end{tabular}
      \end{sc}
    \end{small}
  \end{center}
\end{table}

\noindent\textbf{Subject Coverage.}
Problems are categorized into five broad mathematical subjects: algebra, geometry, number theory, combinatorics, and other.
The benchmark is intentionally balanced across core mathematical domains, with combinatorics, number theory, and algebra each accounting for roughly one quarter of the dataset, as shown in table~\ref{tab:subject_dist}.

\begin{table}[b]
  \caption{Subject distribution of HM-ReasoningBench.}
  \label{tab:subject_dist}
  \begin{center}
    \begin{small}
      \begin{sc}
        \begin{tabular}{lrr}
          \toprule
          Subject & Count & Percentage \\
          \midrule
          Combinatorics & 1188 & 24.1\% \\
          Number Theory & 1167 & 23.7\% \\
          Algebra       & 1164 & 23.6\% \\
          Geometry      & 994  & 20.2\% \\
          Mixed Topics  & 419  & 8.5\% \\
          \bottomrule
        \end{tabular}
      \end{sc}
    \end{small}
  \end{center}
\end{table}

\noindent\textbf{Source Characteristics.}
The two sources exhibit complementary structural properties.
HARP primarily contributes high-difficulty problems, with a strong concentration in Levels~6--7, reflecting its emphasis on advanced multi-step reasoning.
In contrast, OmniMATH spans a broader difficulty spectrum from Level~1 to Level~8 and provides wide subject coverage.
This combination enables HM-ReasoningBench to support both fine-grained difficulty analysis and robust evaluation of reasoning generalization across problem styles.

\noindent\textbf{Intended Use.}
Overall, HM-ReasoningBench is designed to support fine-grained analysis of mathematical reasoning behaviors across subjects, difficulty regimes, and problem styles.
The unified difficulty re-annotation and balanced subject coverage make the benchmark particularly suitable for studying reasoning strategies, cross-domain generalization, and human–model reasoning differences.

\subsection{Strategy Category List}
\label{app:category}

We organize extracted strategies into a fixed set of fine-grained \emph{strategy templates} (i.e., categories), each representing a distinct, recurring reasoning operation (e.g., \texttt{angle\_chasing}, \texttt{modular\_arithmetic}, \texttt{case\_analysis}).
For presentation and aggregation, these templates are further grouped into five broad \emph{subjects}—\emph{Algebraic}, \emph{Number Theory}, \emph{Geometry}, \emph{Combinatorial}, and \emph{Structural}—but all analysis in this paper is conducted at the category level unless stated otherwise.

The complete template list with brief descriptions is provided in Table~\ref{tab:strategy_templates}.

\begin{table*}[t]
  \caption{
    Strategy taxonomy used throughout the paper.
    Extracted strategies are mapped to fine-grained templates capturing
    distinct reasoning operations.
  }
  \label{tab:strategy_templates}
  \begin{center}
    \begingroup
    \small
    \setlength{\tabcolsep}{4pt}
    \renewcommand{\arraystretch}{0.95}
    \begin{tabular}{p{0.16\textwidth}p{0.24\textwidth}p{0.52\textwidth}}
      \toprule
      Subject & Template & Description \\
      \midrule

 \multirow{6}{*}{Algebraic}
& algebraic\_general
& General symbolic manipulation not covered by specialized algebraic templates. \\
& inequality
& Inequality-based reasoning via bounding, convexity, or classical inequalities. \\
& polynomial\_analysis
& Polynomial structure analysis (factorization, roots--coefficients relations, divisibility). \\
& algebraic\_manipulation
& Canonical algebraic transformations (substitution, expansion, identity rewriting). \\
& functional\_equation
& Functional equations and recursive functional constraints. \\
& symmetric\_sum
& Symmetric polynomial arguments and symmetric-sum identities. \\

      \midrule
      \multirow{4}{*}{Number Theory}
      & modular\_arithmetic
      & Modular reasoning and congruence-based arguments. \\
      & prime\_factorization
      & Reasoning via prime decomposition and exponent structure. \\
      & divisibility
      & Divisibility properties and factor-based constraints. \\
      & gcd\_lcm
      & GCD/LCM structure and coprimality arguments. \\

      \midrule
      \multirow{10}{*}{Geometry}
& geometric\_general
& General geometric reasoning not covered by specialized geometric templates. \\
& angle\_chasing
& Angle relations derived from geometric theorems and configurations. \\
& circle\_properties
& Circle geometry (cyclicity, tangency, power of a point, radical axis). \\
& similarity\_congruence
& Similarity or congruence transformations preserving ratios or lengths. \\
& symmetry\_analysis
& Exploiting geometric symmetry to simplify structure. \\
& auxiliary\_construction
& Introducing auxiliary points, lines, or circles to expose hidden relations. \\
& coordinate\_general
& Coordinate-based or analytic reasoning without an explicit coordinate setup. \\
& coordinate\_setup
& Explicit coordinate or analytic setup converting geometry into algebraic constraints. \\
& vector\_method
& Vector-based geometric reasoning (dot/cross products, vector decomposition). \\
& complex\_number
& Complex-plane representations of geometric transformations. \\

      \midrule
      \multirow{5}{*}{Combinatorial}
      & counting\_principle
      & Direct counting arguments (product/sum rules, recurrences). \\
      & inclusion\_exclusion
      & Inclusion--exclusion principle for overlapping sets. \\
      & probability\_method
      & Probabilistic reasoning using probability or expectation. \\
      & bijection
      & Establishing bijections to prove counting equivalences. \\
      & pigeonhole
      & Pigeonhole principle and its generalized forms. \\

      \midrule
      \multirow{5}{*}{Structural}
      & extremal\_principle
      & Extremal arguments via minimal or maximal elements. \\
      & case\_analysis
      & Structured case partitioning and exhaustive enumeration. \\
      & invariant
      & Invariant or monovariant reasoning under transformations. \\
      & proof\_by\_contradiction
      & Contradiction-based arguments assuming negation of the claim. \\
      & mathematical\_induction
      & Inductive reasoning over integers or recursive structures. \\

      \bottomrule
    \end{tabular}
    \endgroup
  \end{center}
\end{table*}

\noindent\textbf{Examples of strategy realizations.}
To make the template descriptions in Table~\ref{tab:strategy_templates} more concrete,
Table~\ref{tab:strategy_examples} lists representative strategy examples.

\begin{table*}[t]
  \caption{
    Representative realizations of strategy templates.
    Each row provides a neutral action description illustrating how a template
    may be instantiated in solutions. These examples are for interpretability
    only and do not affect the taxonomy or experiments.
  }
  \label{tab:strategy_examples}
  \begin{center}
    \begingroup
    \small
    \setlength{\tabcolsep}{4pt}
    \renewcommand{\arraystretch}{0.95}
    \begin{tabular}{p{0.16\textwidth}p{0.24\textwidth}p{0.52\textwidth}}
      \toprule
      Subject & Template & Representative strategy \\
      \midrule

      \multirow{3}{*}{Algebraic}
      & algebraic\_general
      & Write down the given quantities and translate all relationships into
        equations, then solve for the target variable. \\
      & algebraic\_manipulation
      & Introduce intermediate variables to simplify expressions or factor a
        constraint into a product form, then analyze solutions. \\
      & polynomial\_analysis
      & Use coefficient extraction (with index shifts) to derive constraints on
        the coefficient sequence. \\

      \midrule
      \multirow{3}{*}{Combinatorial}
      & counting\_principle
      & Count objects via a direct combination argument (e.g., choose vertices
        and apply a closed-form expression). \\
      & bijection
      & Analyze injectivity or surjectivity of a mapping and account for
        collisions by characterizing exceptional patterns. \\
      & counting\_principle
      & Correct overcounting caused by symmetry or indistinguishable elements
        by dividing by the number of equivalent permutations. \\

      \midrule
      \multirow{3}{*}{Geometry}
      & coordinate\_setup
      & Choose a coordinate system and express geometric constraints as
        algebraic equations, then compute the required quantity. \\
      & auxiliary\_construction
      & Add auxiliary points or lines to expose hidden angle relations or
        similarity structures. \\
      & symmetry\_analysis
      & Place the configuration symmetrically (when permitted) to reduce degrees
        of freedom and simplify the computation. \\

      \midrule
      \multirow{3}{*}{Number Theory}
      & prime\_factorization
      & Restrict candidates via prime factorizations and test feasible exponent
        patterns systematically. \\
      & modular\_arithmetic
      & Establish necessity and sufficiency using congruences, then verify
        remaining cases directly. \\
      & modular\_arithmetic
      & Apply modular constraints to eliminate impossible prime factors before
        checking exceptional cases. \\

      \midrule
      \multirow{3}{*}{Structural}
      & case\_analysis
      & Partition the problem into cases (e.g., parity or magnitude regimes) and
        analyze each case exhaustively. \\
      & mathematical\_induction
      & Conjecture a closed form from initial cases and prove it by induction
        using the recurrence relation. \\
      & proof\_by\_contradiction
      & Assume feasibility and derive a contradiction by comparing magnitudes or
        signs under the given constraints. \\

      \bottomrule
    \end{tabular}
    \endgroup
  \end{center}
\end{table*}

\subsection{Implementation Details of Selective Strategy Retrieval}
\label{app:pool_implementation}

This appendix describes the concrete implementation of Selective Strategy Retrieval (SSR),
including route-specific candidate selection and ranking.
All configurations are fixed across experiments and are not tuned per dataset or model.

\noindent\textbf{Overview.}
SSR retrieves candidate strategies through three routes defined in the main text:
Category-Conditioned Retrieval (Route A),
Problem-Transfer Retrieval (Route B),
and Semantic Fallback Retrieval (Route C).
The final candidate pool is formed by taking the union of strategies retrieved from all routes,
followed by route-aware ranking.

\noindent\textbf{Route A: Category-Conditioned Retrieval.}
SSR first identifies a small set of compatible strategy categories $\mathcal{C}(x)$ for the target problem.
We do \emph{not} train a separate category classifier.
Instead, category compatibility is inferred in the same learned graph embedding space used by SSR:
each category corresponds to a dedicated node in $\mathcal{G}$, and the graph encoder produces embeddings
for problem nodes and category nodes jointly (Appendix~\ref{app:gnn-config}).

At test time, given a problem embedding $h_x$, we score each category node $c\in V_c$ by cosine similarity
$\mathrm{sim}(h_x, h_c)$ and select the top-$2$ categories:
\[
\mathcal{C}(x)=\text{Top2}_{c\in V_c}\ \mathrm{sim}(h_x, h_c).
\]
We then retrieve up to $10$ strategies per selected category based on their similarity to $h_x$ within that
category, forming a compact set of category-consistent candidates.

\noindent\textbf{Route B: Problem-Transfer Retrieval.}
SSR retrieves strategies that were empirically effective on problems in the neighborhood $\mathcal{N}(x)$.
We consider the top $5$ most similar problems
and collect strategies associated with successful guidance on these problems.
This route typically yields a small number of high-precision candidates.

\noindent\textbf{Route C: Semantic Fallback Retrieval.}
When Routes A and B yield insufficient candidates, SSR retrieves additional strategies via semantic similarity.
We perform nearest-neighbor search over \emph{strategy node embeddings} produced by the graph encoder
(Appendix~\ref{app:gnn-config}), using the problem embedding $h_x$ as the query.
We retrieve up to $20$ strategies.
This route is used conservatively and serves only as a fallback.

\noindent\textbf{Candidate Pool Construction.}
Let $\mathcal{S}_A(x)$, $\mathcal{S}_B(x)$, and $\mathcal{S}_C(x)$ denote the strategies retrieved by Routes A, B, and C.
The final candidate pool is constructed as
\[
\mathcal{S}(x) = \mathcal{S}_A(x) \cup \mathcal{S}_B(x) \cup \mathcal{S}_C(x),
\]
with duplicate strategies merged.
\subsection{Executability-Supervised Graph Representation Learning}
\label{app:gnn-config}

To support executability-aware strategy selection, we learn structure-aware node
representations over the strategy knowledge graph $\mathcal{G}$ using supervised
contrastive learning.
\textbf{This module is not used to estimate executability scores or to directly rank
strategies.}
Instead, it provides relational features that are later consumed by the supervised
executability predictor described in Section~\ref{subsec:executability_model}.

\noindent\textbf{Graph construction.}
The heterogeneous graph $\mathcal{G} = (V, E)$ contains three node types:
\textbf{problems} ($V_p$), \textbf{strategies} ($V_s$), and \textbf{categories} ($V_c$).
Edges encode (i) observed problem--strategy associations extracted from correct solutions
in the \emph{training split} of HM-ReasoningBench, and (ii) strategy--category membership.
No information from the evaluation split is used in graph construction or supervision,
ensuring that all executability signals are strictly confined to training data.

\noindent\textbf{Executability supervision.}
We obtain supervision from strategy-guided executions on the training split.
For each evaluated pair $(x,s)$, we run the target model under a fixed protocol for $T$
independent trials and record outcomes $y_{x,s,1:T}\in\{0,1\}$.
We compute a calibrated executability estimate $\tilde U(s\mid x)$ via the Beta--Binomial
posterior mean in Eq.~\eqref{eq:beta_binomial}.
Pairs with $\tilde U(s\mid x)\ge \delta$ are treated as positives; pairs with
$\tilde U(s\mid x)\le \delta^{-}$ are treated as negatives (we fix $\delta=0.5$ and
$\delta^{-}=0.1$ in all experiments), and ambiguous pairs are excluded from contrastive
training.
Unless otherwise stated, we use $T=3$ independent decoding trials per $(x,s)$, and sample
up to $K=10$ negatives per positive pair.

\noindent\textbf{Text encoder for node initialization.}
We initialize \textbf{problem} nodes and \textbf{strategy} nodes with 384-dimensional sentence embeddings
from a pretrained SentenceTransformer encoder (we use \texttt{all-MiniLM-L6-v2} in all experiments).
Category nodes are initialized by mean-pooling the embeddings of strategies assigned to the category.
These initial text features are then refined by the graph encoder via message passing.

\noindent\textbf{Contrastive objective.}
For each positive pair $(x,s^+)$, we sample negatives $\mathcal{N}(x)$ from strategies in
the same category as $s^+$ that are labeled negative for $x$ (falling back to a global
negative pool if necessary).
We optimize the InfoNCE loss:

\begin{equation}
\label{eq:infonce}
\begin{aligned}
\mathcal{L}_{\text{InfoNCE}}
&=
- \sum_{(x,s^+)}
\log
\frac{
\exp\!\left(\mathrm{sim}(h_x, h_{s^+}) / \tau\right)
}{
\exp\!\left(\mathrm{sim}(h_x, h_{s^+}) / \tau\right)
+
\sum_{s^- \in \mathcal{N}(x)}
\exp\!\left(\mathrm{sim}(h_x, h_{s^-}) / \tau\right)
}.
\end{aligned}
\end{equation}
\noindent where $\mathrm{sim}(\cdot,\cdot)$ denotes cosine similarity and $\tau$ is a fixed temperature
hyperparameter ($\tau=0.07$).
This objective encourages executable problem--strategy pairs to be closer in representation
space than non-executable pairs, while controlling for category-level confounds.

\noindent\textbf{Model architecture.}
We use a heterogeneous graph neural network with transformer-based message passing
(\texttt{TransformerConv}).
Separate input projections are applied for each node type (problem, strategy, category),
mapping 384-dimensional SentenceTransformer embeddings into a shared hidden space.
The network consists of three stacked graph transformer layers with four attention heads,
hidden dimension 128, and dropout rate 0.1.
Residual connections and layer normalization are applied after each layer.

\noindent\textbf{Training protocol.}
The graph encoder is trained for 50 epochs using the Adam optimizer with learning rate
$10^{-3}$ and batch size 32.
All hyperparameters are fixed across datasets and target models.
The resulting embeddings are used solely as \emph{structural features} for downstream
executability prediction, and are not directly used to score or select strategies.

\noindent\textbf{Sanity check.}
To verify that the learned representations capture executability-relevant structure,
we evaluate their ability to discriminate executable from non-executable problem--strategy
pairs on a held-out subset of the training split.
Embedding similarity achieves substantially higher AUC than random baselines, indicating
that the contrastive objective encodes meaningful executability information.

\section{Prompt design}
\label{app:prompt}

\subsection{Strategy Extraction Prompt}
\label{app:sa_prompt}

This prompt instructs the model to abstract reusable, high-level problem-solving
strategies from a given worked solution, focusing on transferable reasoning patterns
rather than problem-specific calculations.

\begin{AIBox}{Strategy Extraction Prompt}
\begin{minipage}[t]{\textwidth}
\begin{RaggedRight}
\begin{lstlisting}[basicstyle=\small\ttfamily, breaklines=true, breakatwhitespace=true]
You are a mathematics expert analyzing problem-solving strategies.

Output ONLY valid JSON. Do NOT use markdown or code fences.

Task: extract_solution_strategies

Instructions:
1. From the given solution, extract the KEY STRATEGIES that would help solve
   SIMILAR problems.
2. Write each strategy as a concrete, actionable approach, e.g.:
   - "Express each variable as $p^alpha$ and compare exponents"
   - "Apply inclusion-exclusion to count overlapping cases"

Return the result in the following JSON format:
{
  "strategies": ["strategy1", "strategy2", "..."],
}

Guidelines:
- Limit to 3-5 most critical strategies.
- Focus on reusable reasoning techniques, not problem-specific calculations.
- Be precise, concise, and actionable.

Problem:
{problem_text}

Solution:
{solution_text}
\end{lstlisting}
\end{RaggedRight}
\end{minipage}
\end{AIBox}

\subsection{Direct Answer Prompt}
\label{app:da_prompt}

This prompt serves as a baseline reasoning setup, asking the model to solve a problem
directly without external strategy guidance or example-based hints.

\begin{AIBox}{Direct Answer Prompt}
\begin{minipage}[t]{\textwidth}
\begin{RaggedRight}
\begin{lstlisting}[basicstyle=\small\ttfamily, breaklines=true, breakatwhitespace=true]
Solve the following mathematical problem step by step.

Problem:
{problem_text}

Instructions:
- Provide a detailed solution with clear reasoning.
- Conclude with the final answer.

\end{lstlisting}
\end{RaggedRight}
\end{minipage}
\end{AIBox}

\subsection{Strategy Guidance Prompt}
\label{app:sg_prompt}

This prompt evaluates the effect of explicit strategy-level guidance by providing the
model with strategies extracted from similar problems and instructing it to use them
during solution construction.

\begin{AIBox}{Strategy Guidance Prompt}
\begin{minipage}[t]{\textwidth}
\begin{RaggedRight}
\begin{lstlisting}[basicstyle=\small\ttfamily, breaklines=true, breakatwhitespace=true]
System: You are an expert mathematician solving competition problems.

User: Solve the following problem using the provided strategy guidance.

Problem:
{problem_text}

Strategy guidance (from similar solved problems):
- {strategy_1}
- {strategy_2}
- ...

Instructions:
- Use the strategies above as hints for your solution approach.
- Solve the problem step by step with clear reasoning.
- Conclude with the final answer.
\end{lstlisting}
\end{RaggedRight}
\end{minipage}
\end{AIBox}

\subsection{Answer Verification Prompt}
\label{app:av_prompt}

This prompt is used to automatically assess the correctness of a model-generated answer
by checking mathematical equivalence against a reference solution under strict criteria.

\begin{AIBox}{Answer Verification Prompt}
\begin{minipage}[t]{\textwidth}
\begin{RaggedRight}
\begin{lstlisting}[basicstyle=\small\ttfamily, breaklines=true, breakatwhitespace=true]
System: You are a rigorous mathematics expert who evaluates student solutions with strict standards.

User: You are a rigorous mathematics expert evaluating student answers.

Problem:
{problem_text}

Ground Truth Answer (extracted from reference solution):
{reference_solution}

Student's Final Answer (reasoning process has been separated):
{student_final_answer}

Evaluation task:
Compare the student's final answer with the ground truth answer and determine
whether they are mathematically equivalent.

Mathematical equivalence rules:
- Account for different valid representations:
  * Fractions vs decimals (e.g., 1/2 = 0.5)
  * Mixed numbers vs improper fractions (e.g., 10 2/3 = 32/3)
  * Simplified vs unsimplified forms (e.g., 2/4 = 1/2)
  * Different algebraic forms (e.g., x^2 - 1 = (x-1)(x+1))
  * Equivalent expressions (e.g., 2x + 2 = 2(x+1))
- For proofs: check whether the conclusion is logically equivalent.
- Be STRICT about numerical values (mismatch => wrong).
- Be STRICT about signs (negative vs positive matters).

Classification:
- "Completely Correct": final answer is mathematically equivalent to ground truth
- "Completely Wrong": not equivalent, or missing, or nonsensical

Output ONLY valid JSON (no markdown, no code fences) in this format:
{
  "category": "Completely Correct" | "Completely Wrong",
  "is_correct": true | false,
  "score": 0-100,
  "explanation": "brief explanation"
}
\end{lstlisting}
\end{RaggedRight}
\end{minipage}
\end{AIBox}

\subsection{Strategy Adherence Verification Prompt}
\label{app:adherence_prompt}

This prompt evaluates whether a specific target strategy was actually used—and correctly
executed—in a given reasoning trace, enabling fine-grained analysis of strategy
executability.

\begin{AIBox}{Strategy Adherence Verification Prompt}
\begin{minipage}[t]{\textwidth}
\begin{RaggedRight}
\begin{lstlisting}[basicstyle=\small\ttfamily, breaklines=true, breakatwhitespace=true]
You are an expert evaluator of mathematical reasoning strategies.

Task: Determine whether a specific strategy was correctly executed in a student's solution.

Problem: {problem_text}

Target Strategy:
Strategy Description: {strategy_text}

Student's Reasoning: {reasoning_excerpt}

Evaluation Instructions:
Assess whether the target strategy was actually used in the student's reasoning,
and if so, whether it was applied correctly.

You must classify the strategy usage into exactly one of the following categories:

1. correctly_executed
   - The strategy was genuinely applied
   - The logical steps follow the methodology of the strategy
   - The execution is correct and contributes to the solution

2. attempted_but_incorrect
   - The student attempted to apply the strategy
   - However, the execution contains logical errors, misapplication, or flawed reasoning

3. mentioned_only
   - The strategy is referenced or hinted at via keywords or superficial mentions
   - No substantive application or execution is present

4. not_used
   - There is no evidence that this strategy was used at all

Critical Evaluation Criteria:
- Do NOT rely on keyword matching alone
- Verify that the reasoning steps structurally align with the strategy
- "Correctly executed" requires both correct methodology and correct execution
- Be strict: partial or vague usage should NOT be marked as correctly executed

Output Format:
Return ONLY valid JSON (no markdown, no code fences) in the following format:

{
  "execution_status": "correctly_executed | attempted_but_incorrect | mentioned_only | not_used",
  "confidence": 0-100,
  "evidence": "Direct quote from the student's reasoning, or null if not used",
  "explanation": "2-3 sentences explaining the judgment",
  "critical_to_solution": true | false,
  "execution_quality_score": 0-10
}

Important:
Only mark a strategy as "correctly_executed" if there is clear, concrete evidence
that the student followed the intended strategy and applied it correctly.
\end{lstlisting}
\end{RaggedRight}
\end{minipage}
\end{AIBox}

\section{Trace-Level Case Studies of Strategy Executability}
\label{app:case_studies}

\begin{figure*}[t]
  \vskip 0.1in
  \begin{center}
    \centerline{\includegraphics[width=\linewidth]{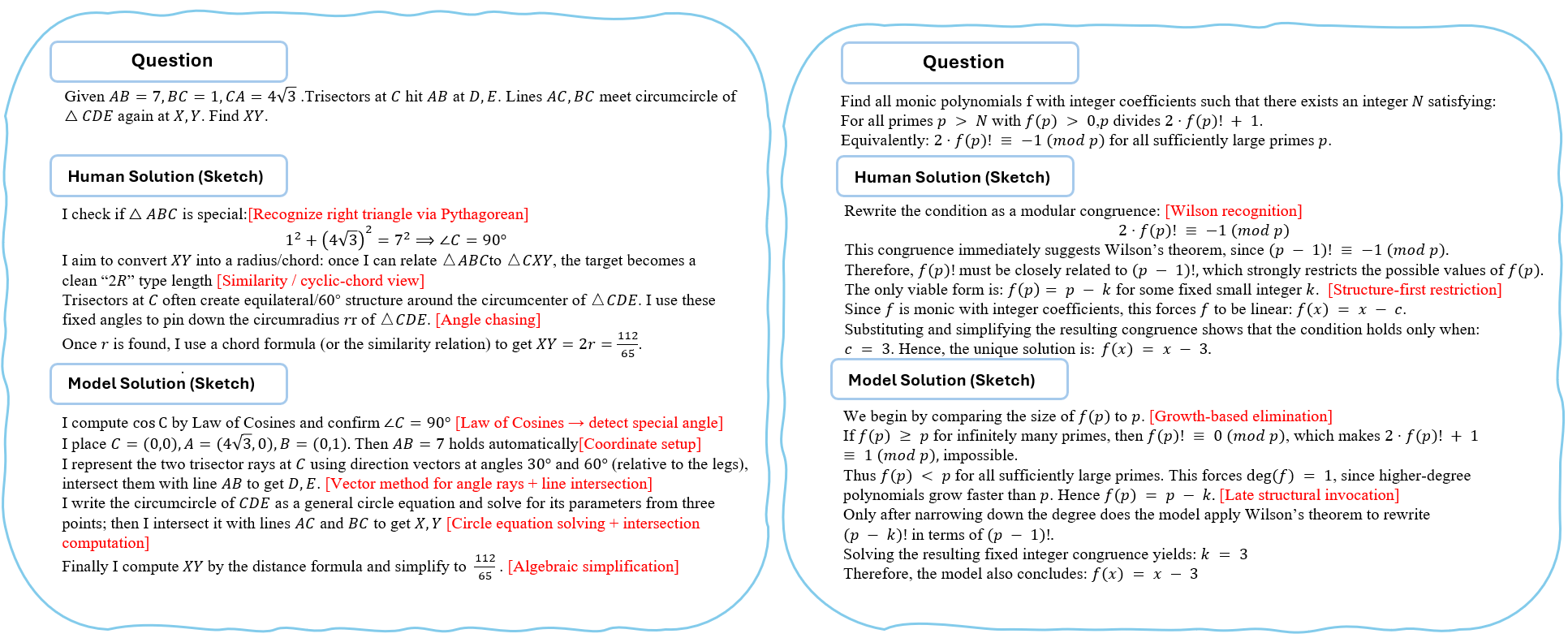}}
    \caption{
      Trace-level case studies illustrating strategy executability differences
      between human-written and model-generated solutions.
      For each problem, we contrast (i) human-derived strategies, which emphasize
      structural recognition or theorem-level reasoning, and (ii) model-derived
      strategies, which rely on procedural or algebraic transformations.
      Although both solution sources reach the correct final answer, the extracted
      strategies exhibit different executability properties.
    }
    \label{fig:trace_case_studies}
  \end{center}
\end{figure*}

Figure~\ref{fig:trace_case_studies} provides concrete trace-level illustrations of the strategy divergences analyzed in Section~\ref{sec:analysis}.
Each example corresponds to a single problem, for which both a human-written solution and a model-generated solution are available and correct.

For each solution, we show the high-level strategies extracted by our pipeline, rather than full step-by-step reasoning.
These examples highlight two recurring phenomena observed throughout our analysis.
First, human-derived strategies often prioritize early structural recognition (e.g., identifying special geometric configurations or invoking strong theorems),
which can be concise but difficult for smaller reasoning models to execute reliably when used as guidance.
Second, model-derived strategies tend to emphasize procedural transformations (e.g., coordinate setups or algebraic elimination),
which are often more executable but may lack global structure or lead to inefficient reasoning when used alone.

\section{More Experiments}

\subsection{Comparison with Stronger Inference-Time Baselines}
\label{app:additional_baselines}

We compare SSR against several inference-time baselines that improve reasoning by
allocating additional test-time computation.
All methods use the same base prompt and model backbone as SSR.

For \textbf{Self-Consistency (SC)}, we sample $N=8$ reasoning traces with non-zero
temperature and apply majority voting over final answers.

For \textbf{Tree-of-Thoughts (ToT)}, we employ a shallow search tree to control inference
cost.
At each step, the model proposes up to $B=3$ candidate continuations, and the search is
truncated to a maximum depth of $D=2$.
Candidate nodes are scored using a lightweight self-evaluation prompt, where the same
model estimates whether a partial reasoning trajectory is likely to lead to a correct
solution.
At each level, only the top-scoring continuation is expanded further, resulting in at
most $1 + B + B$ model calls per problem.

\textbf{Least-to-Most Prompting (L2M)} follows the standard decomposition-and-solve
procedure described in prior work, where the model first decomposes the original problem
into a sequence of simpler subproblems, solves them sequentially, and then composes the
final answer.

Table~\ref{tab:additional_baselines} reports the results of these inference-time baselines.
Across datasets, SC and L2M provide modest improvements over direct solving on
short- and medium-horizon benchmarks.
Tree-of-Thoughts yields stronger improvements than SC and L2M under a bounded compute
budget, yet remains less stable than SSR, particularly on long-horizon problems.
These results indicate that allocating additional test-time computation alone is
insufficient; instead, effective guidance depends on selecting strategies that remain
executable for the target model.

Notably, SSR achieves these improvements using a single guided generation per problem.
In contrast, inference-time baselines such as self-consistency and Tree-of-Thoughts
require multiple model calls to sample or search over reasoning trajectories.
This suggests that gains from SSR do not arise from increased test-time computation,
but from providing strategies that are more executable for the target model.

\begin{table}[h]
  \caption{
    Accuracy (\%) comparison with inference-time baselines.
    All methods use the same base prompt and model backbone.
    Self-Consistency (SC) uses majority voting over $N{=}8$ samples.
    Tree-of-Thoughts (ToT) uses branching factor $B{=}3$ and depth $D{=}2$.
  }
  \label{tab:additional_baselines}
  \begin{center}
    \scriptsize
    \begin{sc}
      \begin{tabular}{lcccc|cccc|cccc}
        \toprule
        & \multicolumn{4}{c}{HM-ReasoningBench}
        & \multicolumn{4}{c}{AIME25}
        & \multicolumn{4}{c}{Apex} \\
        \cmidrule(lr){2-5}
        \cmidrule(lr){6-9}
        \cmidrule(lr){10-13}
        Model
        & SC@8 & L2M & ToT & SSR
        & SC@8 & L2M & ToT & SSR
        & SC@8 & L2M & ToT & SSR \\
        \midrule
        Qwen3-8B
        & 66.40 & 65.60 & 67.20 & \textbf{68.60}
        & 72.00 & 69.33 & 72.33 & \textbf{74.00}
        & 11.42 & 10.61 & 12.24 & \textbf{13.06} \\
        Qwen3-14B
        & 68.80 & 67.40 & 69.60 & \textbf{70.20}
        & 74.33 & 73.33 & \textbf{75.00} & 74.67
        & 13.88 & 11.83 & \textbf{15.91} & 14.69 \\
        R1-Distill-7B
        & 50.60 & 48.80 & 51.20 & \textbf{52.40}
        & 48.67 & 45.33 & 51.33 & \textbf{53.13}
        & 8.97 & 7.35 & \textbf{9.79} & 7.75 \\
        \bottomrule
      \end{tabular}
    \end{sc}
  \end{center}
  \vskip -0.1in
\end{table}

\subsection{Retrieval-Based Baselines Beyond ICL}
\label{app:retrieval_baselines}

The primary baseline table already includes ICL with a semantically retrieved worked
example as a retrieval-based baseline.
To more directly compare SSR against retrieval-style alternatives at the \emph{strategy}
level, we additionally evaluate several end-to-end retrieval baselines on AIME25;
the resulting accuracy comparison is reported in Table~\ref{tab:retrieval_main}.
Each method uses its own candidate construction and ranking strategy while keeping the
same prompt template, hint budget, and decoding protocol as SSR.
BM25 retrieves candidates by lexical similarity, Dense retrieval uses embedding-based
similarity, and MMR applies diversity-aware semantic reranking before guidance
construction.

\subsection{Prior Sensitivity of Executability Estimation}
\label{app:prior_sensitivity}

We use a symmetric Beta prior with $\alpha=\beta=1$ in the main paper.
To verify that results are not an artifact of this choice, we vary only the prior in
Eq.~\eqref{eq:beta_binomial} while keeping the data, target model, decoding protocol,
and retrieval pipeline fixed.

\begin{table}[h]
  \caption{Sensitivity to the Beta prior in Eq.~\eqref{eq:beta_binomial} (Qwen3-14B).}
  \label{tab:prior_sensitivity}
  \begin{center}
    \small
    \begin{tabular}{lccccc}
      \toprule
      Prior $(\alpha,\beta)$ & Spearman vs. $(1,1)$ & Top-20 overlap & HM-RB & AIME25 & Apex \\
      \midrule
      $(0.5,0.5)$ & 0.972 & 93.0\% & 70.00 & 74.33 & 14.69 \\
      $(1,1)$     & 1.000 & 100.0\% & 70.20 & 74.67 & 14.69 \\
      $(2,2)$     & 0.963 & 91.0\% & 69.80 & 74.33 & 14.69 \\
      \bottomrule
    \end{tabular}
  \end{center}
  \vskip -0.1in
\end{table}

The induced rankings remain highly stable, and downstream accuracy varies only slightly,
supporting that the usage--executability findings and the SSR gains are not artifacts of
a particular weak prior.

\subsection{Full Cost Breakdown and Deployment Trade-Offs}
\label{app:cost_breakdown}

Figure~\ref{fig:token_consumption} in the main paper compares \emph{online generation}
tokens only.
Here we add a fuller cost breakdown that includes offline preprocessing/training and
online retrieval overhead.
We use the same Qwen3-14B setup as in the main paper and compare against ToT with the
same bounded-search configuration as Table~\ref{tab:additional_baselines}
($B{=}3$, $D{=}2$).

\begin{table}[h]
  \caption{Average cost per query, averaged across HM-RB, AIME25, and Apex.}
  \label{tab:cost_breakdown}
  \begin{center}
    \small
    \begin{tabular}{lccc}
      \toprule
      Method & Offline cost (GPU-h) & Online latency (s) & Completion tokens \\
      \midrule
      DS & 0.0 & 7.2 & 16286 \\
      ToT ($B{=}3,D{=}2$) & 0.0 & 27.8 & 35715 \\
      SSR & 6.4 & 8.1 & 14114 \\
      \bottomrule
    \end{tabular}
  \end{center}
  \vskip -0.1in
\end{table}

For SSR, the online latency further decomposes into retrieval ($0.11$s), graph query
($0.05$s), and guided generation ($7.94$s).
Thus, SSR has nonzero offline cost, but its additional online non-generation overhead is
small relative to generation, and its per-query online burden is substantially lower
than ToT.
Importantly, the offline stage does \emph{not} fine-tune the base reasoning LLM:
the trainable SSR modules contain about $0.65$M parameters in total (roughly $0.005\%$
of a 14B base model), and there is no gradient update on the base model itself.

\subsection{Feature-Shaping vs. Final Executability Prediction}
\label{app:predictor_ablation}

The graph encoder and the final executability predictor play distinct roles.
To verify that contrastive representation learning contributes beyond the logistic
predictor alone, we compare a version without contrastive graph representation learning
against the full SSR system.

\begin{table}[h]
  \caption{Ablation of contrastive representation learning (Qwen3-14B, accuracy \%).}
  \label{tab:predictor_ablation}
  \begin{center}
    \small
    \begin{tabular}{lccc}
      \toprule
      Setting & HM-RB & AIME25 & Apex \\
      \midrule
      w/o contrastive representation learning & 68.60 & 70.33 & 13.06 \\
      \textbf{Full SSR} & \textbf{70.20} & \textbf{74.67} & \textbf{14.69} \\
      \bottomrule
    \end{tabular}
  \end{center}
  \vskip -0.1in
\end{table}

This confirms that contrastive graph learning and supervised executability prediction are
distinct but complementary: the graph encoder improves structure-aware features, while
the final predictor performs calibrated executability ranking.

\subsection{Task-Conditional Structural vs. Procedural Guidance}
\label{app:style_pool}

To probe the task-conditional interpretation from Section~\ref{subsec:human_analysis},
we run a controlled style-pool analysis in which retrieval is restricted to either a
structural strategy pool or a procedural strategy pool.

\begin{table}[h]
  \caption{Controlled style-pool comparison (accuracy \%).}
  \label{tab:style_pool}
  \begin{center}
    \small
    \begin{tabular}{lccc}
      \toprule
      Problem family (preferred style comparison) & Qwen3-8B & Qwen3-14B & R1-Distill-7B \\
      \midrule
      Geometry / invariant-driven (Structural vs. Procedural) & 71.67 vs. 67.67 & 74.67 vs. 71.67 & 51.33 vs. 48.67 \\
      Counting / modular / algebraic decomposition (Procedural vs. Structural) & 73.33 vs. 68.67 & 75.00 vs. 71.67 & 53.00 vs. 49.67 \\
      \bottomrule
    \end{tabular}
  \end{center}
  \vskip -0.1in
\end{table}

The results support a clear task-conditional pattern: structural guidance helps more
when progress depends on the right high-level reformulation, whereas procedural guidance
helps more when success depends on explicit intermediate operations.

\subsection{Cross-Model Robustness of the Usage--Executability Gap}
\label{app:cross_model_gap}

To test whether the usage--executability gap is specific to compact open-source models,
we repeat the analysis on stronger GPT-series models under the same protocol family.
The gap metric is the mean absolute discrepancy between normalized strategy usage and
normalized executability over the shared strategy set.

\begin{table}[h]
  \caption{Capability trend of the usage--executability gap.}
  \label{tab:capability_gap}
  \begin{center}
    \small
    \begin{tabular}{lcc}
      \toprule
      Model & Gap metric & 95\% CI \\
      \midrule
      R1-Distill-7B & 0.238 & [0.214, 0.263] \\
      Qwen3-8B & 0.221 & [0.199, 0.243] \\
      Qwen3-14B & 0.205 & [0.184, 0.227] \\
      GPT-4.1 & 0.186 & [0.166, 0.206] \\
      GPT-5.1 & 0.173 & [0.154, 0.193] \\
      \bottomrule
    \end{tabular}
  \end{center}
  \vskip -0.1in
\end{table}

The decreasing trend is significant ($p<0.01$), but the gap remains clearly non-trivial
even for stronger models.
High-discrepancy areas remain high-discrepancy as capability increases, including
functional equations ($0.31 \rightarrow 0.27 \rightarrow 0.24$), constructive geometry
($0.29 \rightarrow 0.25 \rightarrow 0.22$), and inequality transformation chains
($0.27 \rightarrow 0.23 \rightarrow 0.21$) for Qwen3-8B $\rightarrow$ GPT-4.1
$\rightarrow$ GPT-5.1.

We also analyze whether executability profiles are structurally consistent across model
families.
Same-series pairs denote Qwen3-8B vs.~Qwen3-14B and GPT-4.1 vs.~GPT-5.1;
heterogeneous pairs are cross-lineage pairs.

\begin{table}[h]
  \caption{Cross-model consistency of executability profiles.}
  \label{tab:cross_model_consistency}
  \begin{center}
    \small
    \begin{tabular}{lcc}
      \toprule
      Metric & Same-series pairs & Heterogeneous pairs \\
      \midrule
      Mean Spearman correlation & 0.84 & 0.76 \\
      Mean overlap@10 & 0.76 & 0.68 \\
      Mean Kendall $\tau$ & 0.66 & 0.57 \\
      \bottomrule
    \end{tabular}
  \end{center}
  \vskip -0.1in
\end{table}

Heterogeneous consistency remains high, indicating a substantial shared cross-model
backbone, while same-series consistency is modestly higher, indicating stable
lineage-specific shifts.
For example, geometric invariant framing is more executable in the GPT series than in
the Qwen series ($0.72$ vs.~$0.58$), illustrating that the shared backbone coexists with
stable model-conditional differences.

\subsection{Robustness of $T=3$ Executability Estimation}
\label{app:t3_robustness}

We perform a higher-trial robustness check on a targeted subset of $(x,s)$ pairs,
including near-threshold cases, comparing the default $T=3$ setting against $T=10$ and
$T=20$.

\begin{table}[h]
  \caption{Robustness of executability estimation under higher trial counts.}
  \label{tab:t3_robustness}
  \begin{center}
    \small
    \begin{tabular}{lcccc}
      \toprule
      Comparison & Spearman $\rho$ & Kendall $\tau$ & Flip rate & $\Delta$ downstream EM \\
      \midrule
      $T=3$ vs. $T=10$ & 0.91 & 0.79 & 5.8\% & +0.28 \\
      $T=3$ vs. $T=20$ & 0.89 & 0.76 & 7.1\% & +0.36 \\
      \bottomrule
    \end{tabular}
  \end{center}
  \vskip -0.1in
\end{table}

Rank agreement remains high, binary flips are limited, and moving to much larger trial
counts yields only small downstream gains.
This supports the use of $T=3$ as a lightweight but stable executability estimate for
the current SSR pipeline.

\subsection{Qualitative Examples}
\label{app:qual_examples}

\noindent\textbf{Example 1: \texttt{OmniMATH\_2281} (Geometry, Level 2).}
\noindent\textbf{Problem.} A square is inscribed in a circle of radius $1$. Find the perimeter of the square.

\medskip
\noindent\textbf{Retrieved Strategies.}
\begin{itemize}
  \item \textbf{(H)} \texttt{circle\_properties}: Use circle radius/diameter relations (e.g., diagonal equals diameter).
  \item \textbf{(M)} \texttt{algebraic\_other}: Recall standard formula $P=4s$.
  \item \ldots
\end{itemize}

\medskip
\noindent\textbf{Reasoning Content (simplified).}
\begin{quote}\small
A square inscribed in a circle has its diagonal equal to the circle's diameter.
Since $r=1$, the diameter is $\textcolor{red}{2}$, so the diagonal is $\textcolor{red}{2}$.
For a square, $\textcolor{red}{d=s\sqrt{2}}$, hence $\textcolor{red}{s=\frac{2}{\sqrt{2}}=\sqrt{2}}$.
Therefore, the perimeter is $\textcolor{red}{P=4s=4\sqrt{2}}$.
\ldots
\end{quote}

\noindent\textbf{Content (final answer, simplified).}
\begin{quote}\small
Using $\textcolor{red}{d=2}$ and $\textcolor{red}{d=s\sqrt{2}}$, we get $\textcolor{red}{s=\sqrt{2}}$ and thus
$\textcolor{red}{P=4\sqrt{2}}$.
\end{quote}

\bigskip

\noindent\textbf{Example 2: \texttt{USAJMO\_2013\_4} (Number Theory, Level 6).}
\noindent\textbf{Problem.}
Let $f(n)$ be the number of ordered representations of $n$ as a sum of powers of $2$.
Find the smallest $n>2013$ for which $f(n)$ is odd.

\medskip
\noindent\textbf{Retrieved Strategies.}
\begin{itemize}
  \item \textbf{(H)} \texttt{counting\_principle}: Characterize when $f(n)$ is odd (often $n=2^k-1$).
  \item \textbf{(H)} \texttt{mathematical\_induction}: Prove the oddness characterization by induction on $k$.
  \item \textbf{(H)} \texttt{counting\_principle}: Derive recurrence $f(n)=\sum_{i} f(n-2^i)$.
  \item \textbf{(M)} \texttt{modular\_arithmetic}: Reduce to parity by defining $\textcolor{red}{g(n)=f(n)\bmod 2}$.
  \item \ldots
\end{itemize}

\medskip
\noindent\textbf{Reasoning Content (simplified).}
\begin{quote}\small
Define $\textcolor{red}{g(n)=f(n)\bmod 2}$ and use the recurrence
$\textcolor{red}{g(n)=\sum_{2^k\le n} g(n-2^k)\bmod 2}$.
By computing small cases, we observe $\textcolor{red}{g(n)=1}$ at $n=1,3,7,15,\ldots$,
suggesting $\textcolor{red}{f(n)\ \text{odd} \iff n=2^k-1}$.
The next such number after $2013$ is $\textcolor{red}{2^{11}-1=2047}$.
\ldots
\end{quote}

\noindent\textbf{Content (final answer, simplified).}
\begin{quote}\small
Since $\textcolor{red}{f(n)\ \text{is odd} \iff n=2^k-1}$, the smallest $n>2013$ is
$\textcolor{red}{2^{11}-1=2047}$.
\end{quote}

\bigskip

\noindent\textbf{Example 3: \texttt{OmniMATH\_3827} (Algebra, Level 5).}
\noindent\textbf{Problem.}
Find all functions $f:\mathbb{R}\to\mathbb{R}$ such that
$f(xy)=f(x)f(y)+f(f(x+y))$ for all $x,y\in\mathbb{R}$.

\medskip
\noindent\textbf{Retrieved Strategies.}
\begin{itemize}
  \item \textbf{(H)} \texttt{functional\_equation}: Verify candidate functions by direct substitution.
  \item \textbf{(M)} \texttt{case\_analysis}: Plug in special values ($x=0,1$, $y=0$, etc.) to derive constraints.
  \item \ldots
\end{itemize}

\medskip
\noindent\textbf{Reasoning Content (simplified).}
\begin{quote}\small
Set $\textcolor{red}{x=0}$ to obtain
$\textcolor{red}{f(0)=f(0)f(y)+f(f(y))}$, hence $\textcolor{red}{f(f(y))=c(1-f(y))}$ where $c=f(0)$.
Test constant solutions: $\textcolor{red}{f\equiv 0}$ works.
Assume affine form $\textcolor{red}{f(x)=ax+b}$ and compare coefficients, yielding $\textcolor{red}{a\in\{0,1\}}$,
and the nonzero affine solution $\textcolor{red}{f(x)=x-1}$.
Finally, \textcolor{red}{verify by substitution} that $f(x)=0$ and $f(x)=x-1$ satisfy the equation.
\ldots
\end{quote}

\noindent\textbf{Content (final answer, simplified).}
\begin{quote}\small
The only solutions are $\textcolor{red}{f(x)\equiv 0}$ and $\textcolor{red}{f(x)=x-1}$.
\end{quote}

\subsection{Failure-Mode Analysis}
\label{app:failure_mode}

We decompose incorrect solutions into two broad failure modes:
(i) \emph{algebraic manipulation errors}, where the global solution plan is largely correct
but execution fails due to symbolic or arithmetic mistakes; and
(ii) \emph{structural reasoning failures}, where the solution fails to establish or exploit
the correct global structure, such as missing a key decomposition, invariant, or case split.

\begin{figure*}[t]
  \vskip 0.1in
  \begin{center}
    \centerline{\includegraphics[width=0.85\linewidth]{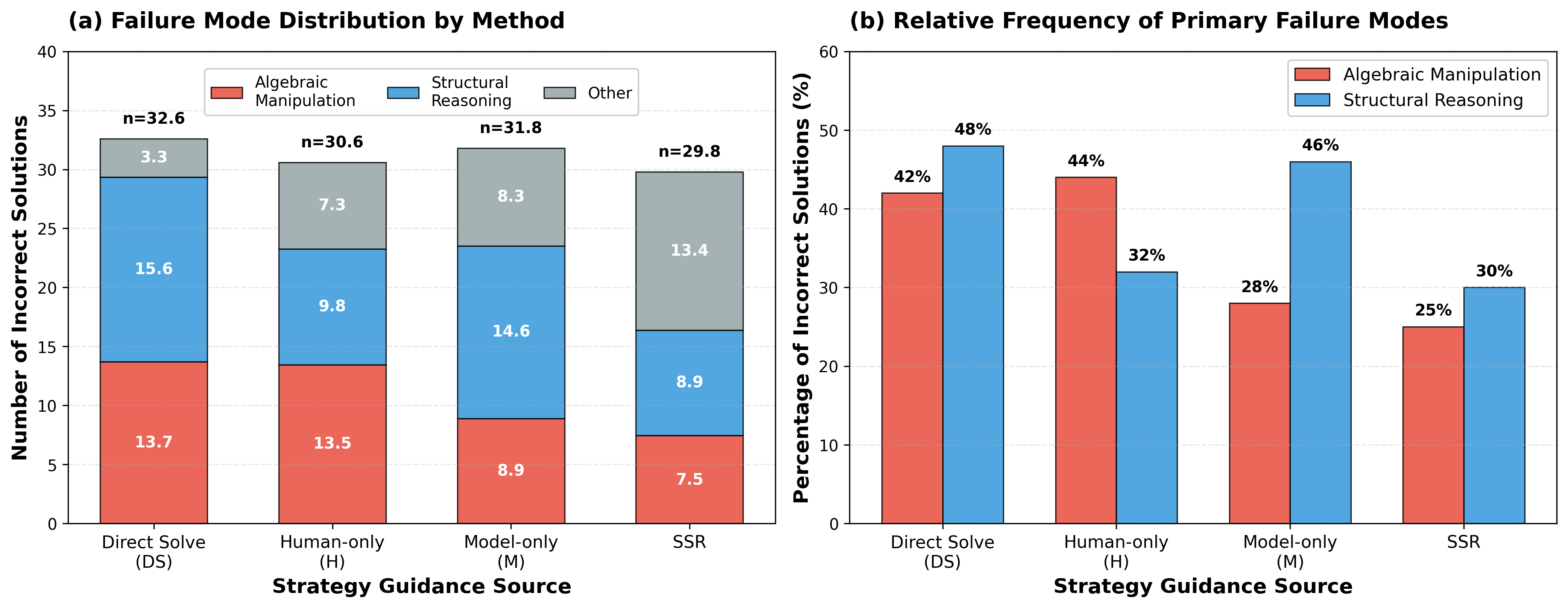}}
    \caption{
      Failure-mode analysis on \textsc{HM-ReasoningBench} using Qwen3-14B.
      Human guidance primarily reduces structural failures, model guidance
      reduces algebraic errors, while SSR mitigates both error types.
    }
    \label{fig:error_type}
  \end{center}
\end{figure*}

Figure~\ref{fig:error_type} reports the distribution of failure modes on
\textsc{HM-ReasoningBench} for Qwen3-14B.
Human guidance primarily reduces structural failures, reflecting its strength in providing
global organization and conceptual structure.
In contrast, model guidance more effectively reduces algebraic manipulation errors.
SSR mitigates both failure modes, consistent with executability-aware selection that
combines complementary structural and procedural signals.

\subsection{Strategy Adherence Evaluation Protocol}
\label{app:adherence}

\begin{figure}[t]
  \vskip 0.1in
  \begin{center}
    \centerline{\includegraphics[width=\linewidth]{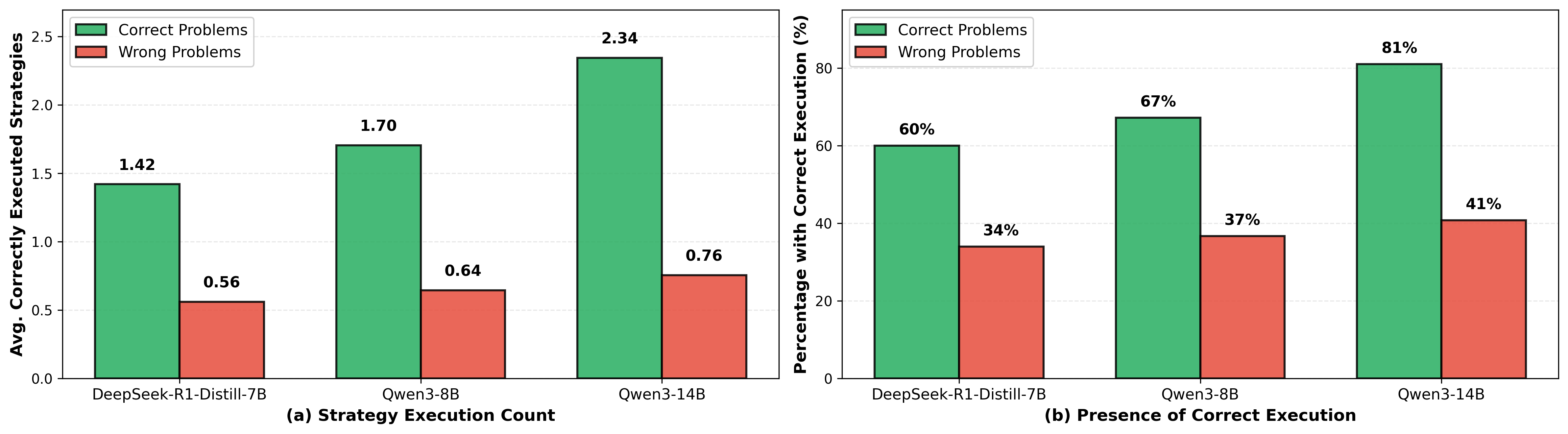}}
    \caption{
      Strategy adherence analysis.
      \textbf{Left:} average number of strategies correctly executed in the
      generated solution.
      \textbf{Right:} percentage of problems for which at least one provided
      strategy is correctly executed.
      Results are reported separately for correct and incorrect final answers.
      Correct solutions consistently exhibit stronger strategy execution, with
      the gap widening for larger models.
    }
    \label{fig:adherence}
  \end{center}
\end{figure}

This appendix provides implementation details for the strategy adherence sanity check
reported in Section~\ref{subsec:adherence}, and summarizes the corresponding results
in Figure~\ref{fig:adherence}.

\noindent\textbf{Setup.}
We randomly sample 100 problems from the HM-ReasoningBench test split and evaluate three
models: DeepSeek-R1-Distill-Qwen-7B, Qwen3-8B, and Qwen3-14B.
For each problem, SSR provides up to five abstract strategy hints as guidance.
Each model generates a single solution under the same prompting and decoding configuration
used in the main experiments.

\noindent\textbf{Adherence criterion.}
We do not require the model to explicitly mention a strategy or follow it verbatim.
A strategy is considered \emph{correctly executed} if the generated solution applies the
strategy in a way that substantively contributes to a valid solution.
Superficial mentions or partial but incorrect applications are not counted as execution.

\noindent\textbf{Evaluation procedure.}
We use GPT-5.1 as an independent evaluator.
The evaluator is provided with (i) the model-generated solution and (ii) the list of
strategy hints given as guidance, and outputs a binary judgment for each strategy
indicating whether it is correctly instantiated in the solution.
The evaluator is instructed to assess functional correctness rather than textual overlap.
The full evaluation prompt is provided in Appendix~\ref{app:adherence_prompt}.

\noindent\textbf{Metrics.}
For each solution, we compute:
(i) the number of correctly executed strategies, and
(ii) a binary indicator of whether at least one strategy is correctly executed.
Statistics are aggregated separately over correct and incorrect final answers.

\noindent\textbf{Results summary.}
Figure~\ref{fig:adherence} reports adherence statistics as a function of model size,
separately for correct and incorrect final answers.
Across models, correct solutions exhibit both a higher number of correctly executed
strategies and a substantially higher probability of executing at least one strategy,
with the gap widening for larger models.

\noindent\textbf{Interpretation.}
Incorrect solutions often exhibit exploratory reasoning that may superficially touch on
multiple strategies without successfully applying any of them.
The proposed metrics therefore focus on whether the provided guidance enables at least one
strategy to be operationalized correctly, directly supporting the executability-based
interpretation discussed in Section~\ref{subsec:adherence}.

\end{document}